\documentclass{article}

\PassOptionsToPackage{numbers}{natbib}

\usepackage[preprint]{neurips_2025}

\usepackage[utf8]{inputenc} 
\usepackage[T1]{fontenc}    
\usepackage{hyperref}       
\usepackage{url}            
\usepackage{booktabs}       
\usepackage{amsfonts}       
\usepackage{nicefrac}       
\usepackage{microtype}      
\usepackage[svgnames,table]{xcolor} 
\usepackage{textcomp}

\usepackage{amsthm,amsmath,amssymb}
\usepackage{mathrsfs}
\usepackage{multirow}
\usepackage{multicol}
\usepackage{color}
\usepackage{hyperref}
\usepackage{caption}
\usepackage{subcaption}
\usepackage{graphicx}

\usepackage{colortbl} 
\usepackage{rotating}
\usepackage{array} 
\usepackage{pifont}
\usepackage{tikz}

\usepackage{wrapfig}
\usepackage{makecell}

\newcommand{\tabincell}[2]{\begin{tabular}{@{}#1@{}}#2\end{tabular}}
\definecolor{LightCyan}{rgb}{0.88,1,1}

\newcolumntype{C}{>{\columncolor{gray!30}}c}
\definecolor{color_light_blue}{HTML}{E0F0FA}
\newcolumntype{H}{>{\setbox0=\hbox\bgroup}c<{\egroup}@{}}
\newcommand{\cmark}{\ding{51}}    
\newcommand{\xmark}{\ding{55}}    

\definecolor{myGreen}{HTML}{75BD42}
\definecolor{myYellow}{HTML}{F2BA02}
\definecolor{myOrange}{HTML}{EE822F}
\definecolor{myRed}{HTML}{FF0000}

\usepackage{enumitem} 

\title{Learning Transferable Facial Emotion Representations from Large-Scale Semantically Rich Captions}

\author{
  Licai~Sun\thanks{Equal contribution.}$~~^1$~~
  Xingxun~Jiang$^{*2}$~~~~
  Haoyu~Chen$^1$~~~~
  Yante~Li$^1$~~~~
  Zheng~Lian$^4$~~~~
  Bin~Liu$^4$~~~~\\
  \textbf{
  Yuan~Zong$^2$~~~~
  Wenming~Zheng\thanks{Corresponding author.}$~~^2$~~~~
  Jukka M.~Leppänen$^3$~~~~
  Guoying~Zhao$\textsuperscript{\dag}~^1$~~~~
  }
  \\
  $^1$University of Oulu\quad
  $^2$Southeast University\quad
  $^3$University of Turku\quad\\
  $^4$Institute of Automation, Chinese Academy of Sciences\quad
  \\
}

\begin{document}
\maketitle

\begin{abstract}
Current facial emotion recognition systems are predominately trained to predict a fixed set of predefined categories or abstract dimensional values. This constrained form of supervision hinders generalization and applicability, as it reduces the rich and nuanced spectrum of emotions into oversimplified labels or scales. In contrast, natural language provides a more flexible, expressive, and interpretable way to represent emotions, offering a much broader source of supervision. Yet, leveraging semantically rich natural language captions as supervisory signals for facial emotion representation learning remains relatively underexplored, primarily due to two key challenges: 1) the lack of large-scale caption datasets with rich emotional semantics, and 2) the absence of effective frameworks tailored to harness such rich supervision. To this end, we introduce EmoCap100K, a large-scale facial emotion caption dataset comprising over 100,000 samples, featuring rich and structured semantic descriptions that capture both global affective states and fine-grained local facial behaviors. Building upon this dataset, we further propose EmoCapCLIP, which incorporates a joint global-local contrastive learning framework enhanced by a cross-modal guided positive mining module. This design facilitates the comprehensive exploitation of multi-level caption information while accommodating semantic similarities between closely related expressions. Extensive evaluations on over 20 benchmarks covering five tasks demonstrate the superior performance of our method, highlighting the promise of learning facial emotion representations from large-scale semantically rich captions. The code and data will be available at \textcolor[rgb]{0.93,0.0,0.47}{\url{https://github.com/sunlicai/EmoCapCLIP}}.

\end{abstract}

\section{Introduction}
\label{sec:intro}

Emotion plays a crucial role in the advancement of artificial intelligence, as it empowers machines to interpret and respond to human affective states, facilitating more natural and harmonious human-machine interactions~\cite{minsky1986society,cowie2001emotion,assunccao2022overview}. 
With the proliferation of deep learning, facial emotion recognition (FER) has emerged as a pivotal research area within the affective computing community, aiming to equip machines with the capability to decode human emotions from facial expressions~\cite{li2020deep,canal2022survey,karnati2023understanding,zhao2023facial}.

\begin{figure*}[t]
    \centering
    \includegraphics[trim={0.0cm 0.73cm 0.0cm 0.0cm},clip,width=\linewidth]{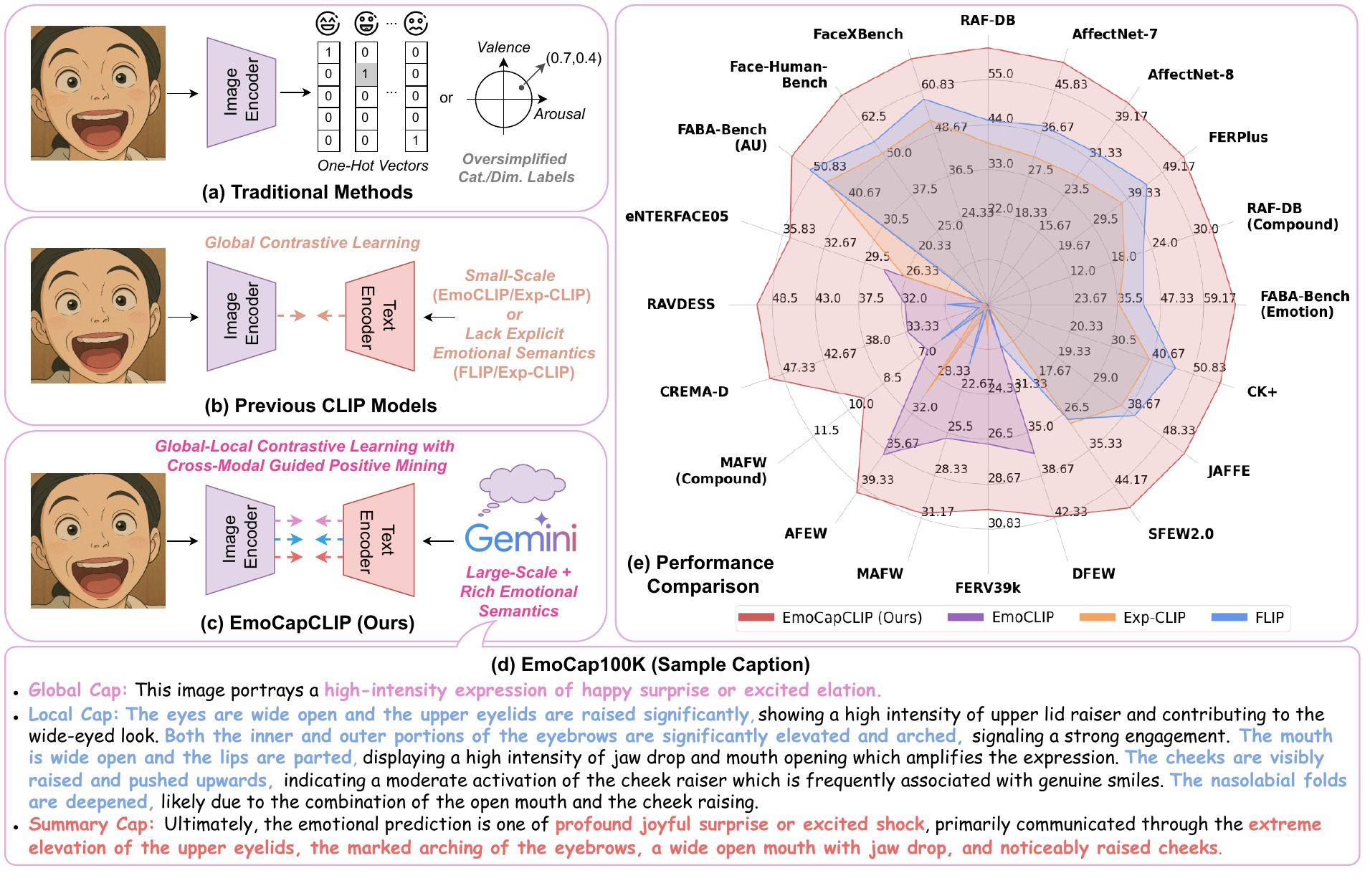}
    \caption{Conceptual overview of EmoCapCLIP. For the performance comparison, we report results based on ViT-B/32. Note: We used GPT-4o~\cite{hurst2024gpt} to convert the original image into Ghibli style.
    }
    \label{fig:teaser}
\end{figure*}

A fundamental challenge in developing FER systems lies in how emotions are represented or annotated. 
Traditionally, two principal paradigms have dominated this landscape: \textit{categorical} method and \textit{dimensional} method (Figure~\ref{fig:teaser}a)~\cite{martinez2012model,rouast2019deep,poria2017review}. 
The categorical method frames emotions as a limited set of categories, such as Ekman's six basic emotions~\cite{ekman1992argument}.
Although this way offers simplicity, it suffers from limited expressiveness, failing to capture emotion intensity or mixed emotions.
The dimensional method, on the other hand, characterizes emotions as numerical coordinates in a multidimensional space~\cite{russell1980circumplex}, allowing for granular modeling of emotional polarity and intensity. 
Despite this capability, it lacks interpretability~\cite{lian2024open} and also struggles to represent compound emotions. 
These limitations inherent in two paradigms constrain FER systems from capturing the rich and nuanced spectrum of human emotions, undermining their generalization and practical applicability.

In contrast, language intrinsically resonates with the complex and nuanced nature of emotional expressions~\cite{cowen2017self,davitz2013language,lindquist2015role,yuan2023describe,li2024facial}, naturally offering a more expressive, flexible, and interpretable alternative for representing and annotating facial emotions. 
Meanwhile, recent advances in vision-language pre-training (e.g., CLIP~\cite{radford2021learning}) have showcased the remarkable power of leveraging language as supervisory signals in general visual understanding.
Nevertheless, the potential of \textit{harnessing semantically rich natural language captions as a broader source of supervision to learn facial emotion representations} has remained relatively untapped in the existing literature (Figure~\ref{fig:teaser}b)~\cite{foteinopoulou2024emoclip,li2024flip,zhao2025enhancing}, mainly due to two major obstacles: 
(1) \textbf{From the data perspective}, existing methods either rely on relatively small, manually annotated datasets with limited diversity and expressiveness~\cite{foteinopoulou2024emoclip}, or utilize generic face captions or latent information that lack explicit emotional semantics~\cite{li2024flip,zhao2025enhancing}. Other available datasets~\cite{li2024facial} are similarly constrained in scale and fail to provide comprehensive, well-structured facial emotion caption that integrates global expressive configurations (e.g., an overall \textit{happily surprised} expression) and corresponding local facial cues (e.g., \textit{highly raised eyebrows} and \textit{wide open mouth}).
(2) \textbf{From the model perspective}, previous efforts predominantly adopt a vanilla global contrastive framework~\cite{foteinopoulou2024emoclip,zhao2025enhancing,li2024flip}, neglecting the modeling of multi-level information embedded in rich emotion captions. Furthermore, semantically related expressions (e.g., \textit{joyful grin} versus \textit{broad smile}) are indiscriminately pushed apart within this framework, which undermines the semantic consistency and generalization ability of the learned emotion representations.

To address the above challenges, we first introduce \textbf{EmoCap100K}, a large-scale semantically
 rich facial emotion caption dataset comprising more than 100,000 samples. 
Given the prohibitive cost and effort of manual annotation, and motivated by the promising performance of multimodal large language models (MLLMs) in emotion-related tasks~\cite{narayan2025facexbench,qin2025face,lian2024gpt},
we employ Gemini-1.5-Flash~\cite{team2024gemini}, a leading proprietary MLLM, for scalable automated caption generation. We leverage carefully designed prompts to elicit comprehensive interpretation of emotional expressions covering both global affective gist and local facial behaviors in a well-structured manner (Figure~\ref{fig:teaser}d). 
Building upon the large-scale EmoCap100K dataset, we further propose \textbf{EmoCapCLIP} to learn emotion-aware facial representations from expressive caption supervision (Figure~\ref{fig:teaser}c). 
Benefiting from the well-structured multi-level information within captions, EmoCapCLIP moves beyond simple global contrast by introducing a joint global-local contrastive learning framework to facilitate more comprehensive and nuanced emotion representation learning.
Moreover, EmoCapCLIP develops a cross-modal guided positive mining module to account for the inherent semantic similarity between closely related emotional expressions, further improving the overall quality of the learned representations.
To verify the effectiveness of our method, we conduct extensive experiments on more than 20 benchmarks covering five FER-related tasks.
The results show that EmoCapCLIP outperforms the state-of-the-art (SOTA) methods by large margins (Figure~\ref{fig:teaser}e). 

\textbf{In summary, our main contributions include:}
(1) We construct EmoCap100K, a large-scale facial emotion caption dataset with comprehensive, well-structured descriptions of global affective gist and local facial behaviors, providing a valuable resource for advancing facial emotion understanding.
(2) We propose EmoCapCLIP, a novel emotion-aware vision-language model equipped with a joint global-local contrastive framework and a cross-modal guided positive mining module for learning rich and nuanced emotion representations.
(3) Extensive evaluations on over 20 benchmarks demonstrate the superior performance of EmoCapCLIP over SOTA methods, validating the effectiveness of learning facial emotion representations from large-scale semantically rich captions.

\section{Related Work}

\subsection{Facial Emotion Recognition}
\textbf{Traditional Paradigms}. With the advent of deep learning, facial emotion recognition (FER) has garnered increasing attention and achieved significant advancements in recent years.
Most existing research in FER focuses on architectural innovations~\cite{li2018occlusion,wang2020region,zhao2021learning,zhao2021robust,ma2021facial,xue2021transfer,zheng2023poster,sun2024svfap,mao2024poster++}) and loss function refinements\cite{li2017reliable,wang2020suppressing,zhang2021relative,zhang2023learning}, aiming to enhance the discriminative power of emotion classifiers. 
However, the supervisory signals employed by these models remain largely constrained to a fixed set of predefined discrete emotion categories (e.g., \textit{happiness, sadness, anger, fear, disgust, surprise, and neutral})~\cite{dhall2012collecting,barsoum2016training,li2017reliable,mollahosseini2017affectnet,jiang2020dfew,liu2022mafw,wang2022ferv39k} or abstract dimensional labels (e.g., \textit{valence, arousal, and dominance})~\cite{mollahosseini2017affectnet,kollias2018aff,kollias2019deep}. 
While these traditional annotation schemes offer advantages in simplicity or granularity, they inherently lack expressiveness, flexibility, and interpretability, thereby constraining the generalization capability and practical applicability of FER systems.

\textbf{Natural Language Supervision.} Natural language, with its inherent semantic depth, provides a more expressive, flexible, and interpretable way to annotate the rich spectrum of emotional expressions. This paradigm aligns deeply with the intrinsic nature of human emotion perception, which is widely acknowledged as continuous (varying in intensity, e.g., \textit{slightly annoyed} and \textit{completely enraged})~\cite{russell1980circumplex}, compound (involving blended affective states, e.g., \textit{happily surprised} and \textit{anxiously excited})~\cite{du2014compound}, and dependent on the synergistic integration of global configurations and local facial cues~\cite{calder2000configural,white2000parts,tanaka2012mixed,krumhuber2023role}.
 Motivated by this advantage and the remarkable success of vision-language pre-training, several recent studies have attempted to leverage language supervision for emotion representations learning~\cite{zhang2023learning,foteinopoulou2024emoclip,zhao2025enhancing}. 
Specifically, EmotionCLIP~\cite{zhang2023learning} learns emotion representations by contrasting verbal and nonverbal cues in daily communications. However, it does not specifically target facial expressions.
To tackle this issue, EmoCLIP~\cite{foteinopoulou2024emoclip} uses short human-annotated emotion captions from MAFW~\cite{liu2022mafw} to perform global contrastive alignment between video and language representations. 
Subsequently, Exp-CLIP~\cite{zhao2025enhancing} adopts an implicit alignment strategy, matching CLIP's visual representations to textual features from BLIP2~\cite{li2023blip}. 
While these methods have shown decent results, they typically depend on small-scale datasets with limited emotional semantics and are constrained by a simplistic global contrast framework, thus failing to fully exploit the potential of semantically rich captions.
In contrast, this paper introduces a large-scale facial emotion caption dataset with rich emotional semantics, and leverages it as the foundation to develop a novel vision-language model that effectively unlocks the power of natural language supervision for facial emotion representation learning.

\subsection{Vision-Language Models}

\textbf{CLIP Models.} Large-scale vision-language pre-training has emerged as a powerful paradigm for learning transferable visual representations~\cite{radford2021learning,jia2021scaling,cherti2023reproducible,zhai2023sigmoid,sun2023eva,xudemystifying,zheng2024dreamlip,lavoie2024modeling,bica2024improving}. 
Pioneering vision-language models such as CLIP~\cite{radford2021learning} and ALIGN~\cite{jia2021scaling} have demonstrated impressive performance across a broad range of computer vision tasks such as image classification, image-text retrieval, and open-vocabulary detection and segmentation.
These models typically employ contrastive learning to align visual and textual modalities on large-scale image-text pairs, resulting in strong zero-shot generalization in downstream tasks. 
Building upon this foundation, recent efforts such as FaRL~\cite{zheng2022general} and FLIP~\cite{li2024flip} have extended vision-language pre-training to the facial domain, aiming to learn general-purpose facial representations.
Despite demonstrating strong performance in various facial analysis tasks (e.g., face alignment and face attribute recognition), they primarily rely on web-crawled generic face captions, which often lack explicit and fine-grained affective semantics, thereby limiting their effectiveness in emotion-centric applications.

\textbf{MLLMs.} 
Beyond CLIP models, recent advances in MLLMs have garnered significant interest, owing to their capacity to jointly understand and reason over diverse modalities~\cite{achiam2023gpt,liu2023visual,zhu2024minigpt,chen2024internvl,bai2023qwen}.
Building on this trend, several studies have explored the development of emotion-centric MLLMs~\cite{li2024facial,lan2025expllm,xing2024emo,cheng2024emotion,lian2025affectgpt,zhao2025r1}, such as EmoLA~\cite{li2024facial}, Emotion-LLaMA~\cite{cheng2024emotion}, and R1-Omni~\cite{zhao2025r1}. 
These models are primarily designed to generate emotion-related captions from image, video, or multimodal inputs, typically through visual instruction tuning.
In contrast, our work takes a fundamentally different perspective: rather than treating caption generation as the end goal, we leverage semantically rich captions as supervisory signals to facilitate the learning of robust facial emotion representations.
Moreover, our approach does not rely on heavy language models and thus has significantly fewer parameters. Nevertheless, it achieves competitive or even superior performance on several benchmarks compared to these substantially larger MLLMs.

\section{EmoCap100K}\label{sec:EmoCap100K}

As shown in Table~\ref{tab:existing_datasets}, our EmoCap100K is distinct from existing face caption datasets. 
\textit{First}, most existing datasets (e.g., LAION-Face~\cite{zheng2022general} and FLIP-80M~\cite{li2024flip}) are constructed for generic facial perception tasks (e.g., facial attributes recognition) rather than being tailored for facial emotion understanding. 
Although they typically contain a large volume of facial images, emotional semantics are largely absent or underrepresented in the associated captions. 
\textit{Second}, MAFW~\cite{liu2022mafw} provides a manually annotated caption dataset dedicated to facial emotions, however, its scale is highly limited, containing less than \textit{one-tenth} of the number of samples in EmoCap100K. Moreover, due to the inherent challenges of manual annotation, the captions are notably short (18 \textit{vs.} 267 words on average) and exhibit limited linguistic diversity in terms of unique emotion words (312 \textit{vs.} 703).
\textit{Third}, the closest counterpart to our dataset is the recently introduced FABA-Instruct~\cite{li2024facial}, which also leverages MLLMs to construct two instruction-tuning datasets focused on facial emotion and action units. Nevertheless, its dataset size is still relatively small, only about \textit{one-fifth} of EmoCap100K. Additionally, the use of simple prompting strategies leads to captions that lack clear structural organization across different semantic components (i.e., global emotional configurations and local facial behaviors), making it difficult to utilize multi-level affective information.

\begin{table*}[hbp]
    \centering
    \caption{Comparison of EmoCap100K with existing face caption datasets.}
    \resizebox{\linewidth}{!}{
    \setlength{\tabcolsep}{1.5pt} 
    \begin{tabular}{lcrrcccc}
       \toprule
       \textbf{Datasets}  & \textbf{\tabincell{c}{Emotion-Centric}} & \textbf{\#Sample} & \textbf{\#Avg. Words} & \textbf{\tabincell{c}{\#Uni.\\Emo. Words}} & 
       \textbf{Structured} & \textbf{Data Source} & \textbf{Caption Source} \\ 
        \midrule
                          
       FFHQ-Text~\cite{zhou2021generative}  &  \xmark  & 760 & 183 &- & \xmark  & Internet  & Human\\
       MM-CelebA-HQ~\cite{xia2021tedigan}  & \xmark & 30,000  & 17 &-   & \xmark & Internet & Human+Template\\

       CelebA-Dialog~\cite{jiang2021talk}  &   \xmark & 202,599 & 25 & -  & \xmark & Internet & Human+Template \\
       LAION-Face \cite{zheng2022general}  & \xmark  & 50,000,000  & 12 & -     & \xmark & Internet & Web-Crawled\\
       FaceCaption-15M \cite{dai202415m}  & \xmark  & 15,000,000  &30 & -  & \xmark & Internet & Model+Template+LLM
       
        \\
       FLIP-80M \cite{li2024flip} & \xmark & 83,000,000 & 16 &-  & \xmark & Internet+AIGC  & Web-Crawled+LLM \\
       \midrule
       MAFW\cite{liu2022mafw}  & \cmark & 8,034  & 18 & 312  &\xmark  & Screen Media & Human \\

       FABA-Instruct (Emotion) \cite{li2024facial} & \cmark & 19,877 & 50 & 650  & \xmark  & Internet  & MLLM\\
       
       FABA-Instruct (AU) \cite{li2024facial}   &\cmark & 16,163 & 207 & 698   &\xmark & Internet & MLLM\\
       \rowcolor{color_light_blue}
       EmoCap100K (Ours)  & \cmark & \textbf{107,134}  & \textbf{267} &\textbf{703}  & \cmark & Screen Media  & MLLM\\
       \bottomrule
    \end{tabular}   
    }
    \label{tab:existing_datasets}
\end{table*}

\subsection{Data Collection}
From a data-centric viewpoint, one promising approach to improve the model's generalization ability is to collect large-scale affective facial data that exhibits spontaneous and diverse emotions in unconstrained, real-world settings~\cite{li2017reliable}.
As a typical type of screen media, movies have been considered as a reliable data source to collect required facial expressions because actors perform in naturalistic conditions and vividly mimic spontaneous facial expressions~\cite{jiang2020dfew,liu2022mafw}. 
Following this rationale, we curated a large-scale dataset by downloading over 1,000 movies spanning various genres on open-source platform and extracting human faces from them.
To ensure diversity in the collected data, we focused on three key aspects: head poses, scene contexts, and emotion categories.
In particular, for emotional diversity, we aimed to go beyond the six basic emotions and include compound and atypical emotional expressions~\cite{kumar2024measuring}.
To maximize the collection of desired facial samples, we established a set of collection rules and recruited multiple volunteers to manually select the data accordingly.
Detailed collection rules and data statistics are provided in Appendix C.

\subsection{Caption Generation}
Research in psychology indicates that human perceiving facial expressions of emotion is a complex process involving the synergistic interplay of global and local information processing~\cite{calder2000configural,white2000parts,adolphs2002recognizing,tanaka2012mixed,krumhuber2023role}. 
Specifically, local processing allows for detailed analysis of individual facial components, while global processing enables the rapid perception of the overall configuration and integration of features into a meaningful whole. 
Following this insight, our aim is to generate captions that offer a comprehensive and well-structured interpretation of facial expressions by integrating both the global expressive configuration and specific local facial behaviors that contribute to this global impression.

\begin{figure}[htbp]
    \centering
    \begin{subcaptionbox}{Emotion Words\label{fig:a}}[0.3\linewidth]
        {\includegraphics[width=\linewidth]{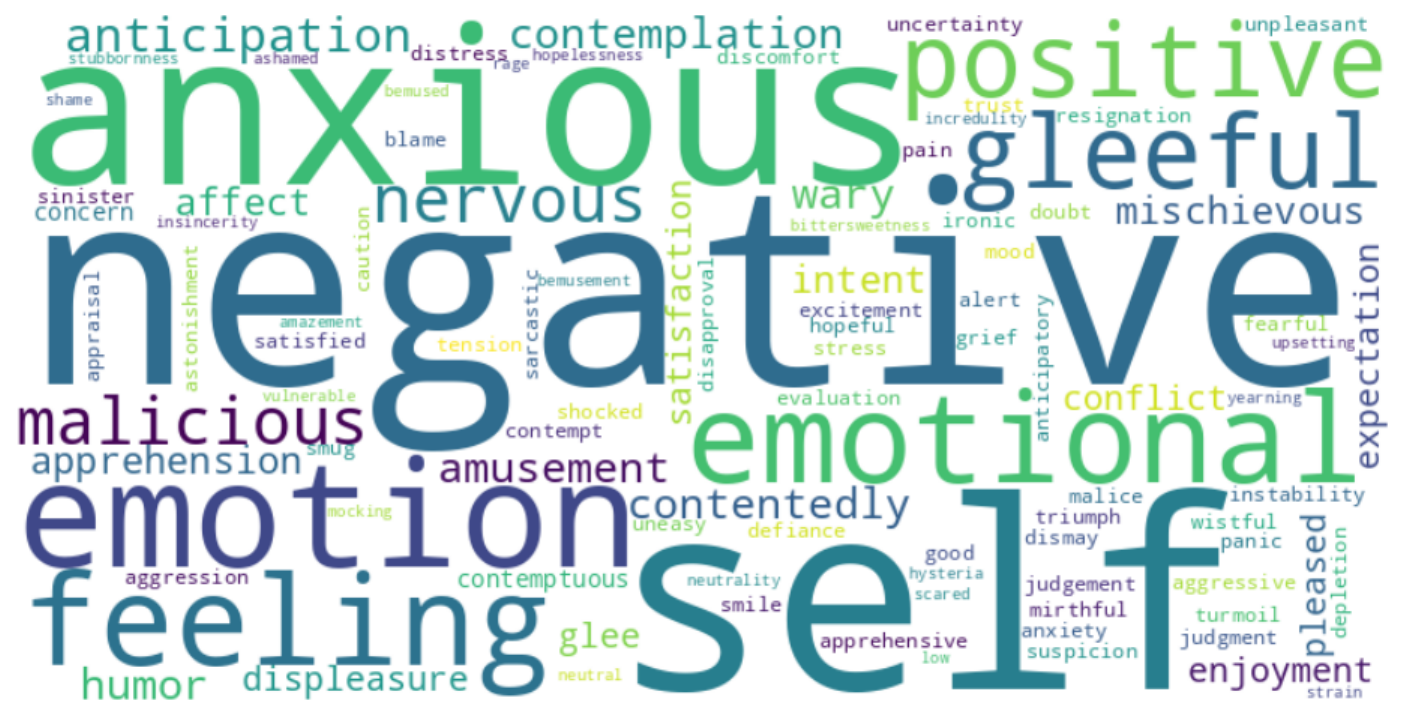}}
    \end{subcaptionbox}
    \hfill
    \begin{subcaptionbox}{Emotional Intensity\label{fig:b}}[0.3\linewidth]
        {\includegraphics[width=\linewidth]{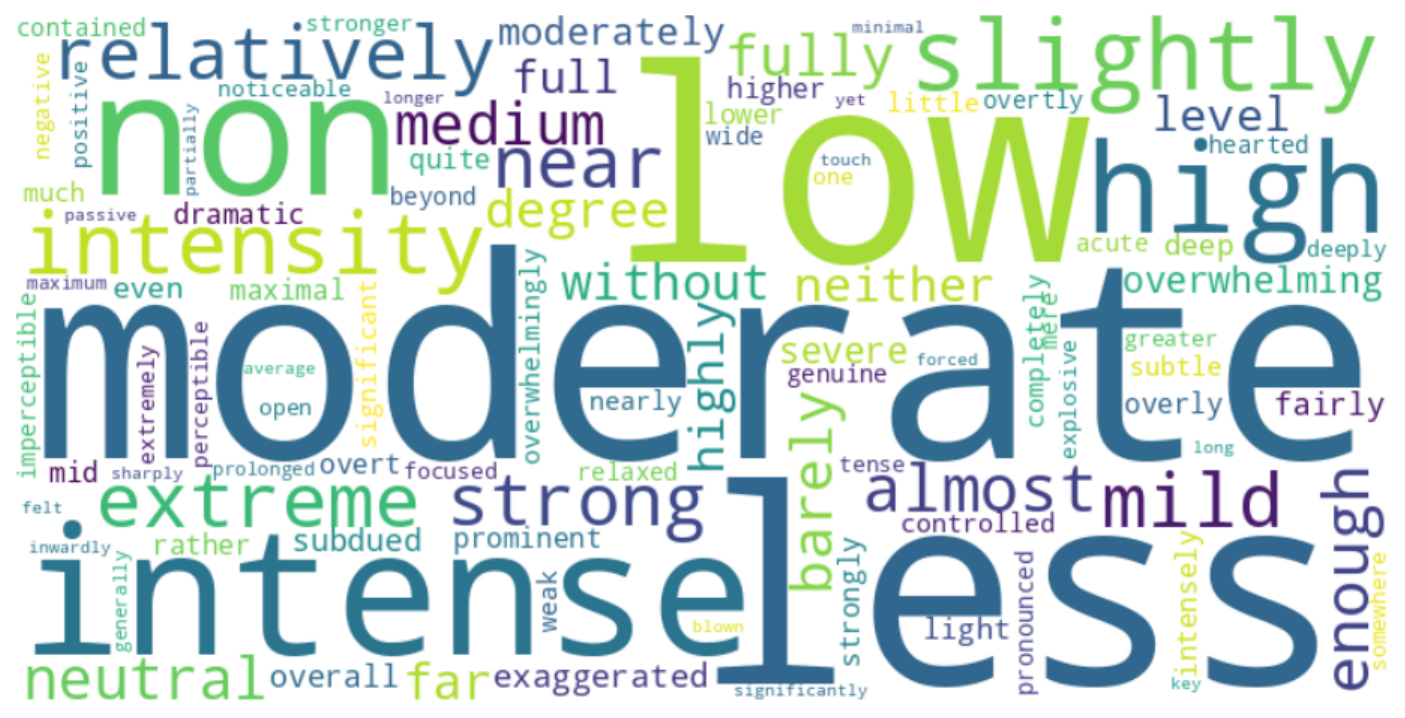}}
    \end{subcaptionbox}
    \hfill
    \begin{subcaptionbox}{Local Emotional Cues\label{fig:c}}[0.3\linewidth]
        {\includegraphics[width=\linewidth]{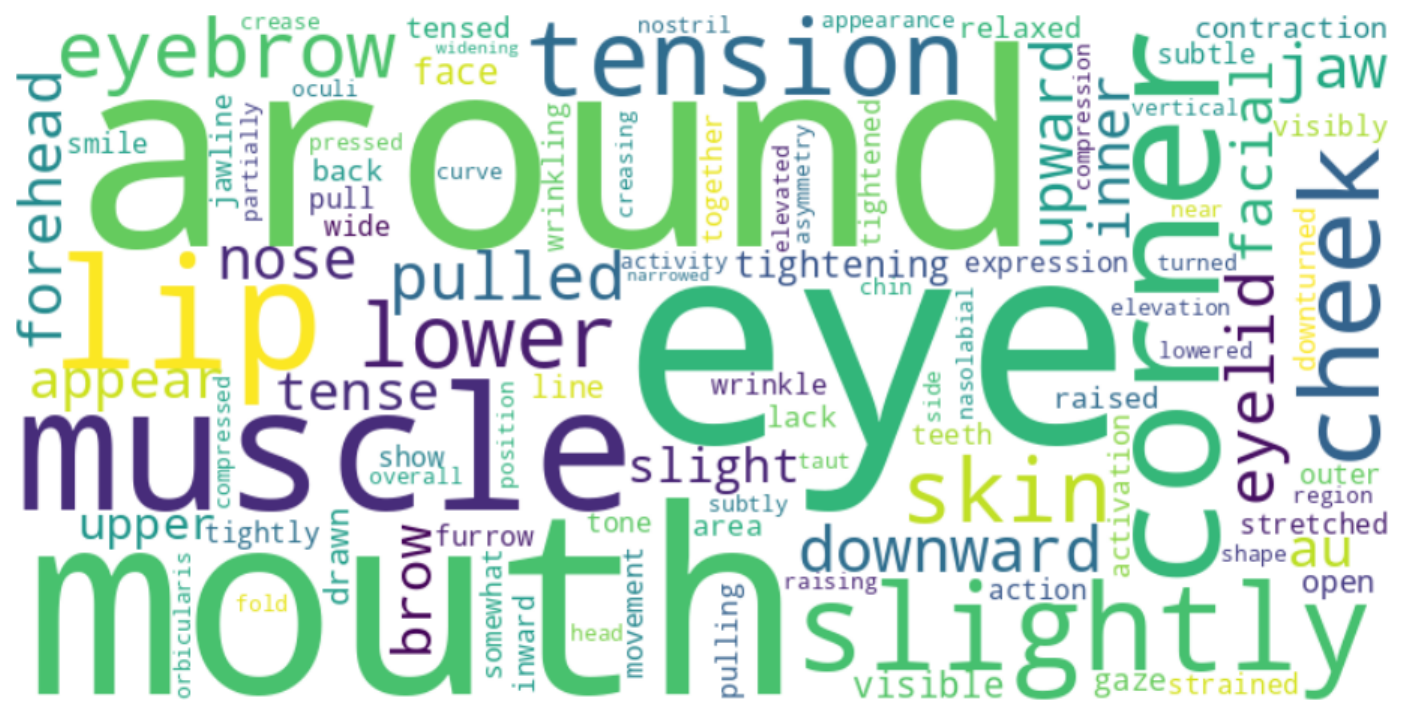}}
    \end{subcaptionbox}
    \caption{Word clouds of EmoCap100K dataset, showing rich words to describe facial emotions.}
    \label{fig:wordclouds-v3.1}
\end{figure}

Generally speaking, manual annotation is considered one of the most reliable labeling methods. However, as mentioned above, manually annotated captions are often short and lack diversity. Moreover, the process is time-consuming and labor-intensive, making it expensive to scale data size.
Fortunately, several recent studies have demonstrated the strong performance of proprietary MLLMs in facial emotion-related tasks~\cite{narayan2025facexbench,qin2025face,lian2024gpt}.
Considering that Gemini-1.5-Flash~\cite{team2024gemini} was a highly capable yet cost-effective LLM at annotating time, we employed it to generate large-scale captions in an automated manner.
Specifically, to effectively elicit the desired captions, we carefully designed a prompt that emphasizes multi-level emotional information across three components: (1) a single sentence describing the global expressive configuration, (2) detailed descriptions of local facial cues that contribute to the global impression, with one sentence devoted to each cue, and (3) an integrative concluding sentence that synthesizes both the global expression and local details into a coherent summary.
The detailed prompt is provided in Appendix C.
As shown in Figure~\ref{fig:teaser}(d), the generated caption is both comprehensive and well-structured, closely aligning with human interpretations of emotional expressions.
Moreover, the three word clouds in Figure~\ref{fig:wordclouds-v3.1} highlight the semantic richness and diversity of the generated captions, further underscoring their descriptive depth.

\section{EmoCapCLIP}
\begin{figure*}[t]
    \centering
    \includegraphics[trim={0.5cm 0.48cm 1.4cm 0.1cm},clip,width=\linewidth]{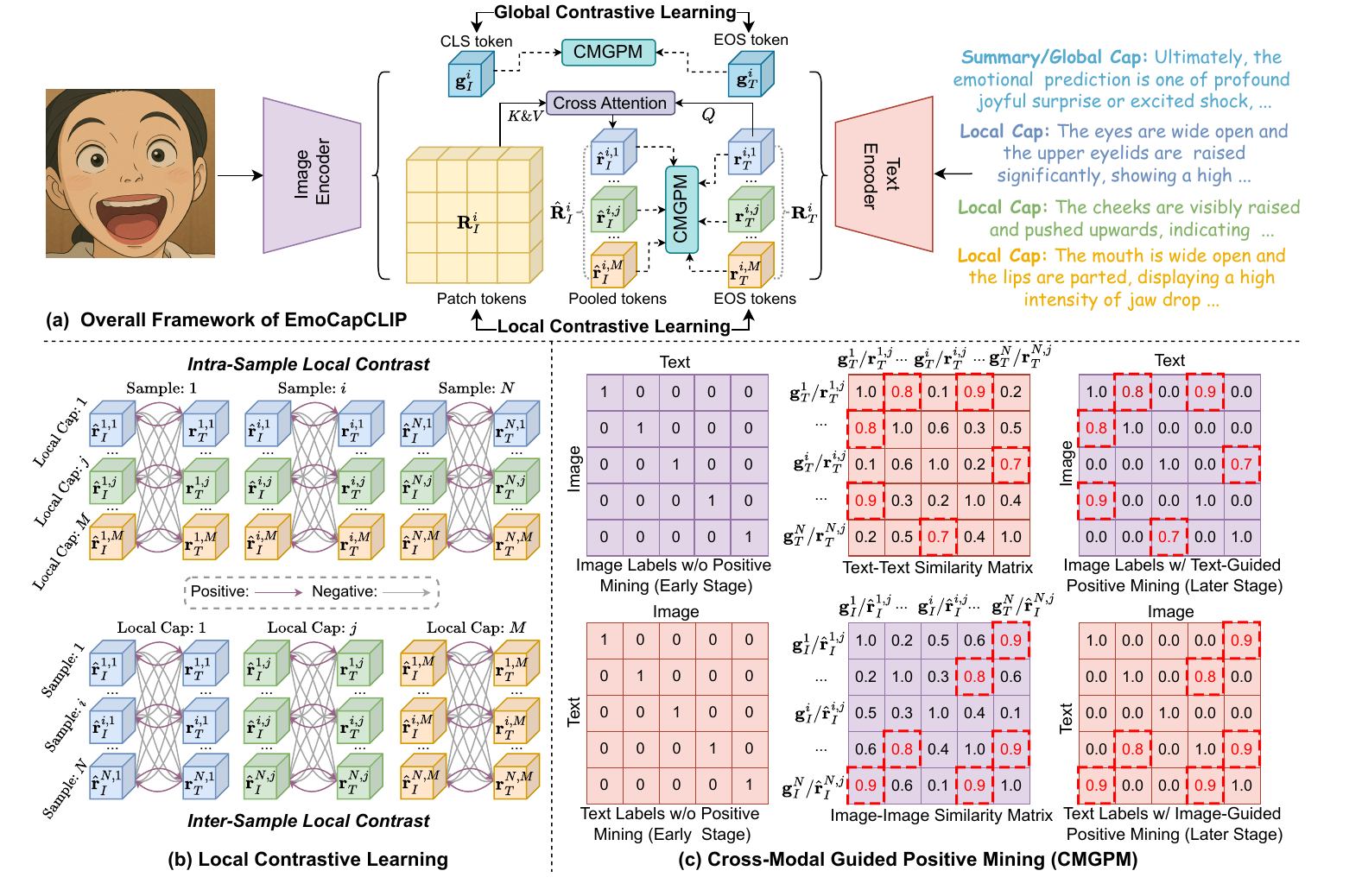}
    \caption{The illustration of EmoCapCLIP, featuring a joint global-local contrastive framework (a-b) and a cross-modal guided positive mining module (c) to fully harness rich supervision in captions.}
    \label{fig:overall_framework}
\end{figure*}

As illustrated in Figure~\ref{fig:overall_framework}(a), EmoCapCLIP primarily features: (i) a joint global-local contrastive learning framework (Section~\ref{sec:jglcl}) that fosters comprehensive emotion understanding by fully exploiting multi-level semantic information embedded in rich, structured captions, and (ii) a cross-modal guided positive mining module (Section~\ref{sec:cmgpm}) aimed at promoting semantic consistency and enhancing contrastive representation learning through the identification of more informative positive samples.

\subsection{Joint Global-Local Contrastive Learning}
\label{sec:jglcl}
EmoCapCLIP follows CLIP~\cite{radford2021learning} to employ a dual-encoder architecture, utilizing a Vision Transformer (ViT~\cite{dosovitskiy2020image}) for image encoding and a Transformer~\cite{vaswani2017attention} for text encoding. 
For convenience in the following discussion, we assume that $\mathcal{B} = \{I^i, T^i\}_{i=1}^N$ is a batch of $N$ image-caption pairs, where the caption includes three parts, i.e., $T^i=[T^i_{\texttt{Global}}, T^i_{\texttt{Local}}, T^i_{\texttt{\texttt{Summary}}}]$.

\textbf{Global Contrastive Learning.}
As shown in Figure~\ref{fig:overall_framework}(a), the image $I^i$ is passed through the image encoder to produce the global image representation $\mathbf{g}_I^i \in \mathbb{R}^{D}$ (derived from the \texttt{[CLS]} token~\cite{dosovitskiy2020image}). 
For the caption $T^i$, since both $T^i_{\texttt{Global}}$ and $T^i_{\texttt{\texttt{Summary}}}$ contain the overall description of affective states, we feed one of them into the text encoder to obtain the global text representation $\mathbf{g}_T^i \in \mathbb{R}^{D}$ (derived from the \texttt{[EOS]} token~\cite{radford2021learning}). 
Following CLIP, especially during the early training stage, for each image (or caption), EmoCapCLIP treats the paired caption (or image) as the positive sample and all other captions (or images) in the batch as negative samples.
This leads to the formulation of the global contrastive loss:
\begin{equation}
\small
\mathcal{L}_g = -\frac{1}{N} \sum_{i=1}^{N} \Bigl[
\log \frac{\exp{(\mathcal{S} \langle \mathbf{g}^{i}_{I}, \mathbf{g}^{i}_{T} \rangle / \tau)}}{\sum_{n=1}^{N} \exp{(\mathcal{S} \langle \mathbf{g}^{i}_{I}, \mathbf{g}^{n}_{T} \rangle / \tau)}} +
\log \frac{\exp{(\mathcal{S} \langle \mathbf{g}^{i}_{T}, \mathbf{g}^{i}_{I} \rangle / \tau)}}{\sum_{n=1}^{N} \exp{(\mathcal{S} \langle \mathbf{g}^{i}_{T}, \mathbf{g}^{n}_{I} \rangle / \tau)}} \Bigr],
\label{eq:loss_global}
\end{equation}
where $ \mathcal{S} \langle \cdot, \cdot \rangle$ denotes the cosine similarity function and $\tau$ is a learnable temperature parameter.

\textbf{Local Contrastive Learning.} 
Local facial behaviors are the building blocks of emotional expressions, encapsulating critical details that characterize specific emotional nuances~\cite{krumhuber2023role,ekman1978facial}. 
Hence, supplementing global contrastive learning with local contrastive learning is essential to achieve a more comprehensive and nuanced understanding of facial emotions. 
To do so, as depicted in Figure~\ref{fig:overall_framework}(a), we first randomly sample $M$ sentences from $T^i_{\texttt{Local}}$ without replacement, each of which details one of local behaviors in specific regions. Then we pass them through the text encoder to obtain their sentence representations $\mathbf{R}_T^{i} = \{ \mathbf{r}_T^{i,j} \}_{j=1}^{M} \in \mathbb{R}^{M \times D}$. 
To facilitate local contrast, a cross-attention layer, whose query is from $\mathbf{R}_T^{i}$ and key/value comes from the dense image patch embeddings $\mathbf{R}_I^i \in \mathbb{R}^{P \times D}$ ($P$ is the number of patch tokens), is employed to obtain the pooled image representations $\mathbf{\hat{R}}_I^{i} = \{ \mathbf{\hat{r}}_I^{i,j} \}_{j=1}^{M} \in \mathbb{R}^{M \times D}$, i.e., 
$\mathbf{\hat{R}}_I^{i} = \textrm{softmax}(\mathbf{Q}^{i}_T {\mathbf{K}_I^{i}}^\top / \sqrt{D}) \mathbf{V}^i_I$, 
where $\mathbf{Q}^{i}_T = \mathbf{R}_T^{i} \mathbf{W}^{Q}$, $\mathbf{K}^{i}_I = \mathbf{R}_I^{i} \mathbf{W}^{K}$, $\mathbf{V}^{i}_I = \mathbf{R}_I^{i} \mathbf{W}^{V}$, and $\mathbf{W} \in \mathbb{R}^{D\times D}$ are projection matrices.

As shown in Figure~\ref{fig:overall_framework}(b), the local contrastive loss operates on the sets of local sentence representations $\mathbf{R}_T^{i} = \{ \mathbf{r}_T^{i,j} \}_{j=1}^{M}$ and the associated pooled image representations $\mathbf{\hat{R}}_I^{i} = \{ \mathbf{\hat{r}}_I^{i,j} \}_{j=1}^{M}$. 
Since the elements in them correspond to distinct facial behaviors in specific regions, it is intuitive that we can treat representations from other local regions of the same sample as negatives and the paired one as the positive~\cite{bica2024improving, zheng2024dreamlip}. 
This formulates the \textit{intra-sample} local contrastive loss: 
\begin{equation}
\small
\begin{split}
\mathcal{L}_{r}^{\textrm{intra}} &= -\frac{1}{NM} \sum_{i=1}^{N} \sum_{j=1}^{M} \Bigl[ 
\log \frac{\exp{(\mathcal{S} \langle \mathbf{\hat{r}}^{i,j}_{I}, \mathbf{r}^{i,j}_{T} \rangle / \tau)}}{\sum_{m=1}^{M} \exp{(\mathcal{S} \langle \mathbf{\hat{r}}^{i,j}_{I}, \mathbf{r}^{i,m}_{T} \rangle / \tau)}} +
\log \frac{\exp{(\mathcal{S} \langle \mathbf{r}^{i,j}_{T}, \mathbf{\hat{r}}^{i,j}_{I} \rangle / \tau)}}{\sum_{m=1}^{M} \exp{(\mathcal{S} \langle \mathbf{r}^{i,j}_{T}, \mathbf{\hat{r}}^{i,m}_{I} \rangle / \tau)}} \Bigr].
\end{split}
\label{eq:loss_local_intra}
\end{equation}
Although minimizing $\mathcal{L}_r^{\textrm{intra}}$ promotes learning discriminative features for different local facial behaviors, the contrastive signal derived solely from within a single sample can be weak due to the limited number of distinct local descriptions (e.g., $M=3$ in our experiments). To address this, we also introduce \textit{inter-sample} local contrastive loss to incorporate more diverse negative samples. Following global contrastive loss $\mathcal{L}_g$, particularly during early training, this loss leverages other samples in the batch as negatives to improve local contrastive representation learning. It is formulated as:
\begin{equation}
\small
\begin{split}
\mathcal{L}_r^{\textrm{inter}} &= -\frac{1}{NM} \sum_{i=1}^{N} \sum_{j=1}^{M} \Bigl[
\log \frac{\exp{(\mathcal{S} \langle \mathbf{\hat{r}}^{i,j}_{I}, \mathbf{r}^{i,j}_{T} \rangle / \tau)}}{\sum_{n=1}^{N} \exp{(\mathcal{S} \langle \mathbf{\hat{r}}^{n,j}_{I}, \mathbf{r}^{i,j}_{T} \rangle / \tau)}} +
\log \frac{\exp{(\mathcal{S} \langle \mathbf{r}^{i,j}_{T}, \mathbf{\hat{r}}^{i,j}_{I} \rangle / \tau)}}{\sum_{n=1}^{N} \exp{(\mathcal{S} \langle \mathbf{r}^{n,j}_{T}, \mathbf{\hat{r}}^{i,j}_{I} \rangle / \tau)}} \Bigr].
\end{split}
\label{eq:loss_local_inter}
\end{equation}

\subsection{Cross-Modal Guided Positive Mining (CMGPM)}
\label{sec:cmgpm}

A strong assumption in the above formulation of $\mathcal{L}_g$ and $\mathcal{L}_r^{\textrm{inter}}$ is that only the explicitly paired representations represent a true positive match, while all other $N-1$ non-paired ones within the batch (i.e., $(\mathbf{g}^{i}_{I}, \mathbf{g}^{n}_{T})$ or $(\mathbf{\hat{r}}^{i,j}_{I}, \mathbf{r}^{n,j}_{T})$ for $n \neq i$) are \textit{complete} mismatches serving as negative samples. 
However, this assumption often fails to hold, especially considering the continuous spectrum of emotions (e.g., non-paired two faces exhibiting \textit{joyful grin} versus \textit{broad smile}) ~\cite{cowen2017self,gao2024softclip,srinivasa2023cwcl}. 
To this end, CMGPM is introduced to identify these inappropriate negatives and treat them as positives, thereby enhancing semantic consistency and improving the overall quality of learned representations.

\textbf{Contrastive Loss with CMGPM.}
As illustrated in Figure~\ref{fig:overall_framework}(c), the core idea of CMGPM is to leverage similarities within one modality as guidance to identify potential positive samples for the other modality. 
Without loss of generality, let us consider the global image representation $\mathbf{g}^{i}_{I}$ as the anchor. The default positive is the paired global text representation $\mathbf{g}^{i}_{T}$.
We posit that if another text representation $\mathbf{g}^{p}_{T}$ in the batch is similar to $\mathbf{g}^{i}_{T}$, then the caption of sample $p$ is likely to be semantically close to that of sample $i$, thus making $\mathbf{g}^{p}_{T}$ a potential positive for $\mathbf{g}^{i}_{I}$.
Therefore, we first calculate the cosine similarity of $\mathbf{g}^{i}_{T}$ with all global text representations (i.e., $\{ \mathbf{g}^{p}_{T} \}_{p=1}^N$). We then identify a positive set $\mathcal{P}_T^i$ for the anchor $\mathbf{g}^{i}_{I}$. This set comprises the sample whose cosine similarity with $\mathbf{g}^{i}_{T}$ is above a predefined threshold $\sigma$ and is among the top $K$ most similar samples in the batch. 
Formally, the positive set $\mathcal{P}_T^i$ for $\mathbf{g}^{i}_{I}$ is defined as:
\begin{equation}
\small
\mathcal{P}_T^i = \{ p \mid \mathcal{S} \langle \mathbf{g}^{i}_{T}, \mathbf{g}^{p}_{T} \rangle > \sigma \text{ and } \mathcal{S} \langle \mathbf{g}^{i}_{T}, \mathbf{g}^{p}_{T} \rangle \in \textrm{topK} \big( \{ \mathcal{S} \langle \mathbf{g}^{i}_{T}, \mathbf{g}^{p'}_{T} \rangle \}_{p'=1}^N \big), p \in [1, \dots, N] \}.
\end{equation}
Note that this set $\mathcal{P}_T^i$ includes the index $i$ itself, as $\mathcal{S} \langle \mathbf{g}^{i}_{T}, \mathbf{g}^{i}_{T} \rangle = 1$. Besides, $\mathcal{P}_T^i$ degenerates into $\{i\}$ (i.e., without positive mining) when $\sigma \to 1$ or $K=1$. 
The positive set $\mathcal{P}_I^i$ for $\mathbf{g}^{i}_{T}$ can be defined similarly, which is omitted it here.
Thus, the global contrastive loss with CMGPM is formulated as:
\begin{equation}
\small
\mathcal{L}_g' = -\frac{1}{N} \sum_{i=1}^{N} \Bigl[
\sum_{p \in \mathcal{P}_T^i} \lambda_{T}^{i,p} \log \frac{\exp{(\mathcal{S} \langle \mathbf{g}^{i}_{I}, \mathbf{g}^{p}_{T} \rangle / \tau)}}{\sum_{n=1}^{N} \exp{(\mathcal{S} \langle \mathbf{g}^{i}_{I}, \mathbf{g}^{n}_{T} \rangle / \tau)}} +
\sum_{p \in \mathcal{P}_I^i} \lambda_{I}^{i,p} \log \frac{\exp{(\mathcal{S} \langle \mathbf{g}^{i}_{T}, \mathbf{g}^{p}_{I} \rangle / \tau)}}{\sum_{n=1}^{N} \exp{(\mathcal{S} \langle \mathbf{g}^{i}_{T},  \mathbf{g}^{n}_{I} \rangle / \tau)}} \Bigr],
\label{eq_loss_global_pm}
\end{equation}
where $\lambda_T^{i,p} = \mathcal{S} \langle \mathbf{g}^{i}_{T}, \mathbf{g}^{p}_{T} \rangle$ and $\lambda_I^{i,p} = \mathcal{S} \langle \mathbf{g}^{i}_{I}, \mathbf{g}^{p}_{I} \rangle$ are weights assigned to the mined positive pairs. 
Finally, the inter-sample local loss with CMGPM (i.e., $\mathcal{L}_r^{\textrm{inter}'}$) is similarly defined and omitted.

\textbf{Overall Training Loss.}
In the early training stage, we adopt the standard formulation of global and inter-sample local contrastive losses without CMGPM, as the representation spaces of both modalities are still immature and unable to capture reliable sample relationships. As training progresses and representations become more stable, we activate the CMGPM module to leverage cross-modal complementary cues for mining high-quality positives, thereby facilitating mutual refinement of representations across modalities. Formally, the overall training loss is defined as:
\begin{equation}
    \mathcal{L}_{\text{overall}} = 
    \begin{cases} 
       \mathcal{L}_{g} + \alpha (\mathcal{L}_{r}^{\textrm{intra}} + \mathcal{L}_{r}^{\textrm{inter}}) 
      & \textrm{if epoch} < t, \\ 
       \mathcal{L}_{g}' + \alpha (\mathcal{L}_{r}^{\textrm{intra}} + \mathcal{L}_{r}^{\textrm{inter}'})
      & \textrm{if epoch} \ge t,
    \end{cases}
    \label{eq:overall_loss}
\end{equation}
where $\alpha$ is the weight for local contrastive loss and $t$ is the epoch at which CMGPM starts.

\section{Experiments}
\label{sec:exp}

\subsection{Main Results}
To comprehensively evaluate EmoCapCLIP, we conduct experiments across five tasks, including \textit{zero-shot static facial expression recognition} (SFER), \textit{zero-shot dynamic facial expression recognition} (DFER), \textit{zero-shot facial action unit detection} (FAUD), \textit{zero-shot expression-caption retrieval} (ECR), and \textit{few-shot} SFER. 
The main evaluation metrics are the unweighted average recall (UAR) and the weighted average recall (WAR, i.e., accuracy). 
Due to space constraints, we only report a subset of the results for zero-shot SFER/DFER and few SFER. 
The implementation \& evaluation details and more results for other tasks are provided in Appendix E,F.

\textbf{Zero-shot SFER}. 
We first evaluate EmoCapCLIP on 9 popular SFER benchmarks. 
To ensure a fair comparison, we follow Exp-CLIP~\cite{zhao2025enhancing} to use the prompt “\textit{a photo of a face with an expression of} [CLASS].” to evaluate all CLIP models. 
As shown in Table~\ref{tab:zero_shot_sfer_no_oulu}, EmoCapCLIP outperforms both general-purpose and domain-specific CLIP models by significant margins on all benchmarks. 
For instance, when using ViT-B/32, EmoCapCLIP beats Exp-CLIP~\cite{zhao2025enhancing}—a method specifically designed for zero-shot SFER—by over 20\% on RAF-DB and $\sim$10\% on RAF-DB (Compound). 
It also substantially outperforms FLIP~\cite{li2024flip}, a strong generic face CLIP model trained on FLIP-80M—800$\times$ larger than EmoCap100K, thereby underscoring the value of semantically rich captions and task-tailored contrastive learning.
Besides, EmoCapCLIP can push performance to even higher levels by scaling up the backbone capacity or compute.  For example, EmoCapCLIP (ViT-L/14) improves EmoCapCLIP (ViT-B/32) by more than 6\%, 9\%, 15\% in terms of UAR on RAF-DB, RAF-DB (Compound), and CK+, respectively.
Furthermore, compared to much larger MLLMs that have billion-level parameters, EmoCapCLIP also demonstrates comparable or even better performance. Specifically, EmoCapCLIP (ViT-L/14) surpasses EmoLA~\cite{li2024facial} by 13\% UAR on FABA-Bench (Emotion) and EMO-LLaMA~\cite{xing2024emo} by 5\% UAR on AffectNet-8. More surprisingly, it slightly outperforms proprietary Gemini-1.5-Flash~\cite{team2024gemini} and GPT-4V~\cite{achiam2023gpt} on several benchmarks (e.g., AffectNet-8 and CK+). 
To sum up, the above encouraging results indicate that EmoCapCLIP can learn powerful facial emotion representations from large-scale semantically rich captions.

To further examine the generalization capability of EmoCapCLIP, we compare its zero-shot performance against the cross-dataset performance of traditional supervised baselines.
Following the previous method~\cite{zhang2024generalizable}, we report the results of WAR using ViT-B/32 in Table~\ref{tab:zero_shot_vs_supervised_cross_eval}. 
The zero-shot performance of EmoCapCLIP is better than the cross-dataset evaluation results of multiple supervised methods (e.g., CAFE~\cite{zhang2024generalizable}, FaRL~\cite{zheng2022general}, and ExpLLM~\cite{lan2025expllm}) on AffectNet-7 and SFEW2.0, even though they are trained with manually annotated labels.
This suggests that our approach has stronger generalization capabilities, highlighting the benefit of leveraging natural language with rich semantics as a much broader source of supervision rather than relying solely on a fixed set of predefined categories.
Although the performance on RAF-DB is relatively lower, it is worth noting that these supervised models are trained on AffectNet-7, which is over $2\times$ larger than EmoCap100K.

\begin{table*}[!t]
\scriptsize
\begin{center}
\caption{Zero-Shot SFER results (UAR/WAR) in comparison with SOTA CLIP models \& MLLMs. Best UAR/WAR in \textbf{bold}, second best \underline{underlined} per section and benchmark.}
\label{tab:zero_shot_sfer_no_oulu}
\resizebox{\linewidth}{!}{
\setlength{\tabcolsep}{1.5pt} 
\begin{tabular}{lHccccccccccc}
\toprule

\textbf{Methods} & \textbf{Data} &  \textbf{\#Param.} &
  \multicolumn{1}{c}{\textbf{RAF-DB}} &
  \multicolumn{1}{c}{\textbf{AffectNet-7}} &
  \multicolumn{1}{c}{\textbf{AffectNet-8}} &
  \multicolumn{1}{c}{\textbf{FERPlus}} &
  \multicolumn{1}{c}{\textbf{\tabincell{c}{RAF-DB\\(Compound)}}} &
  \multicolumn{1}{c}{\textbf{\tabincell{c}{FABA-Bench\\(Emotion)}}} &
  \multicolumn{1}{c}{\textbf{CK+}} &
  \multicolumn{1}{c}{\textbf{JAFFE}} &
                      \textbf{SFEW2.0} \\

\midrule
& \multicolumn{9}{c}{\texttt{MLLMs}} & \\
\midrule

FLAVA~\cite{singh2022flava}  & - &  - &
  \multicolumn{1}{c}{14.35/38.69} &
  \multicolumn{1}{c}{14.26/14.26} &
  \multicolumn{1}{c}{12.47/12.48} &
  \multicolumn{1}{c}{12.50/28.43} &
  \multicolumn{1}{c}{-} &
  \multicolumn{1}{c}{-} &
  \multicolumn{1}{c}{-} &
  \multicolumn{1}{c}{-} &
                     - \\

BLIP2~\cite{li2023blip}   & - &  11B &
  \multicolumn{1}{c}{43.78/44.07} &
  \multicolumn{1}{c}{\underline{32.86}/\underline{32.87}} &
  \multicolumn{1}{c}{28.53/28.53} &
  \multicolumn{1}{c}{43.40/48.14} &
  \multicolumn{1}{c}{-} &
  \multicolumn{1}{c}{-} &
  \multicolumn{1}{c}{-} &
  \multicolumn{1}{c}{-} &
                      - \\

Face-MLLM~\cite{sun2024face}   &  - &  7B &
  \multicolumn{1}{r}{-/67.40} &
  \multicolumn{1}{c}{-} &
  \multicolumn{1}{c}{-} &
  \multicolumn{1}{c}{-} &
  \multicolumn{1}{c}{-} &
  \multicolumn{1}{c}{-} &
  \multicolumn{1}{c}{-} &
  \multicolumn{1}{c}{-} &
                      - \\

EmoLA~\cite{li2024facial}   &  - &  7B &
  \multicolumn{1}{c}{-} &
  \multicolumn{1}{c}{-} &
  \multicolumn{1}{c}{-} &
  \multicolumn{1}{c}{-} &
  \multicolumn{1}{c}{-} &
  \multicolumn{1}{c}{\underline{54.48}/\underline{64.50}} &
  \multicolumn{1}{c}{-} &
  \multicolumn{1}{c}{-} &
                      - \\

EMO-LLaMA~\cite{xing2024emo}   &  - &  7B &
  \multicolumn{1}{r}{-/\underline{71.51}} &
  \multicolumn{1}{c}{-} &
  \multicolumn{1}{r}{38.43/38.43} &
  \multicolumn{1}{r}{-/59.29} &
  \multicolumn{1}{c}{-} &
  \multicolumn{1}{c}{-} &
  \multicolumn{1}{c}{-} &
  \multicolumn{1}{c}{-} &
                      - \\

GPT-4V~\cite{achiam2023gpt,lian2024gpt}  & - &  - &
  \multicolumn{1}{r}{\textbf{68.99}/\textbf{75.81}} &
  \multicolumn{1}{c}{-} &
  \multicolumn{1}{r}{\underline{42.77}/\underline{42.77}} &
  \multicolumn{1}{r}{\underline{48.74}/\underline{64.25}} &
  \multicolumn{1}{c}{-} &
  \multicolumn{1}{c}{-} &
  \multicolumn{1}{r}{\underline{61.06}/\underline{69.72}} &
  \multicolumn{1}{c}{-} &
                     \textbf{50.05}/\textbf{57.24} \\

Gemini-1.5-Flash~\cite{team2024gemini}  & - &  - &
  \multicolumn{1}{r}{\underline{68.25}/70.99} & 
  \multicolumn{1}{c}{\textbf{53.03}/\textbf{53.03}} &
  \multicolumn{1}{r}{\textbf{46.39}/\textbf{46.39}} &
  \multicolumn{1}{r}{\textbf{64.89}/\textbf{66.12}} &
  \multicolumn{1}{c}{\textbf{36.09}/\textbf{43.80}} &
  \multicolumn{1}{c}{\textbf{74.07}/\textbf{73.45}} &
  \multicolumn{1}{r}{\textbf{75.06}/\textbf{83.18}} &
  \multicolumn{1}{c}{\textbf{62.45}/\textbf{62.45}} &
                     \underline{48.46}/\underline{52.44}
\\

\midrule
& \multicolumn{9}{c}{\texttt{CLIP: ViT-B/32}} & 
\\ \midrule

OpenCLIP~\cite{cherti2023reproducible}  & LAION-400M & 149M &
  \multicolumn{1}{c}{31.13/31.23} &
  \multicolumn{1}{c}{26.11/26.11} &
  \multicolumn{1}{c}{19.68/19.68} &
  \multicolumn{1}{c}{29.39/31.18} &
  \multicolumn{1}{c}{18.28/20.45} & 
  \multicolumn{1}{c}{29.18/25.56} & 
  \multicolumn{1}{c}{30.67/35.47} & 
  \multicolumn{1}{c}{24.54/25.35} &
                      21.88/21.11 \\ 

OpenAI CLIP~\cite{radford2021learning} &  WIT-400M & 149M &
  \multicolumn{1}{c}{37.68/30.80} &
  \multicolumn{1}{c}{31.71/31.71} &
  \multicolumn{1}{c}{26.75/26.75} &
  \multicolumn{1}{c}{35.36/43.32} &
  \multicolumn{1}{c}{21.02/17.93} &
  \multicolumn{1}{c}{34.85/44.91} &
  \multicolumn{1}{c}{28.14/38.23} &
  \multicolumn{1}{c}{29.95/29.58} &
                      30.85/30.39 \\

MetaCLIP~\cite{xudemystifying} &  MetaCLIP-400M & 149M &
  \multicolumn{1}{c}{32.41/32.86} &
  \multicolumn{1}{c}{29.49/29.49} &
  \multicolumn{1}{c}{24.28/24.28} &
  \multicolumn{1}{c}{26.93/37.56} &
  \multicolumn{1}{c}{18.38/17.17} &
  \multicolumn{1}{c}{34.07/46.65} &
  \multicolumn{1}{c}{30.27/34.66} &
  \multicolumn{1}{c}{30.47/30.52} &
                      19.59/23.67 \\

SigLIP2~\cite{tschannen2025siglip} &  WebLI & 149M &
  \multicolumn{1}{c}{\underline{49.46}/48.24} &
  \multicolumn{1}{c}{\underline{41.63}/\underline{41.63}} &
  \multicolumn{1}{c}{31.10/31.10} &
  \multicolumn{1}{c}{42.91/47.83} &
  \multicolumn{1}{c}{\underline{25.68}/\underline{26.89}} &
  \multicolumn{1}{c}{\underline{48.83}/\underline{56.08}} &
  \multicolumn{1}{c}{\underline{47.11}/50.36} &
  \multicolumn{1}{c}{\underline{42.40}/\underline{42.72}} &
                      \underline{36.55}/\underline{38.28} \\

Exp-CLIP~\cite{zhao2025enhancing} &  \tabincell{c}{WIT-400M+CAER-S} & 149M &
  \multicolumn{1}{c}{39.56/42.14} &
  \multicolumn{1}{c}{31.73/31.73} &
  \multicolumn{1}{c}{27.63/27.63} &
  \multicolumn{1}{c}{37.83/37.04} &
  \multicolumn{1}{c}{19.98/21.72} &
  \multicolumn{1}{c}{36.03/42.18} &
  \multicolumn{1}{c}{40.09/53.82} &
  \multicolumn{1}{c}{36.99/36.62} &
                     {28.88/32.48} \\

FLIP~\cite{li2024flip}  &   \tabincell{c}{WIT-400M+FLIP-80M} & 149M &
  \multicolumn{1}{r}{45.14/\underline{49.32}} &
  \multicolumn{1}{c}{38.34/38.34} &
  \multicolumn{1}{r}{\underline{32.90}/\underline{32.90}} &
  \multicolumn{1}{c}{\underline{44.76}/\underline{48.47}} &
  \multicolumn{1}{c}{22.76/20.58} &
  \multicolumn{1}{c}{42.72/46.90} &
  \multicolumn{1}{c}{46.46/\underline{55.15}} &
  \multicolumn{1}{c}{40.64/40.85} &
                      27.85/28.54 \\

\rowcolor{color_light_blue}
EmoCapCLIP (Ours) &  \tabincell{c}{WIT-400M+EmoCap-100K} & 149M &
  \multicolumn{1}{c}{\textbf{62.89}/\textbf{67.03}} &
  \multicolumn{1}{c}{\textbf{51.91}/\textbf{51.91}} &
  \multicolumn{1}{c}{\textbf{43.40}/\textbf{43.40}} &
  \multicolumn{1}{c}{\textbf{55.10}/\textbf{61.73}} &
  \multicolumn{1}{c}{\textbf{32.76}/\textbf{30.93}} &
  \multicolumn{1}{c}{\textbf{67.96}/\textbf{68.46}} &
  \multicolumn{1}{c}{\textbf{57.59}/\textbf{65.11}} &
  \multicolumn{1}{c}{\textbf{54.48}/\textbf{54.81}} &
                       \textbf{49.38}/\textbf{53.06} \\

\midrule
& \multicolumn{9}{c}{\texttt{CLIP: ViT-B/16}} &\\

\midrule

OpenCLIP~\cite{cherti2023reproducible}  &   LAION-400M & 149M &
  \multicolumn{1}{c}{36.39/38.62} &
  \multicolumn{1}{c}{26.74/26.74} &
  \multicolumn{1}{c}{21.43/21.43} &
  \multicolumn{1}{c}{33.57/43.32} &
  \multicolumn{1}{c}{20.07/24.12} &
  \multicolumn{1}{c}{32.02/41.94} &
  \multicolumn{1}{c}{24.05/22.73} &
  \multicolumn{1}{c}{33.81/33.80} &
                      28.02/34.80\\

OpenAI CLIP~\cite{radford2021learning}  &   WIT-400M & 149M &
  \multicolumn{1}{c}{40.48/35.79} &
  \multicolumn{1}{c}{37.06/37.06} &
  \multicolumn{1}{c}{31.67/31.67} &
  \multicolumn{1}{c}{40.05/43.32} &
  \multicolumn{1}{c}{22.11/23.11} &
  \multicolumn{1}{c}{37.29/39.95} &
  \multicolumn{1}{c}{46.29/47.71} &
  \multicolumn{1}{c}{25.18/24.88} &
                      25.63/25.52 \\

MetaCLIP~\cite{xudemystifying} &   MetaCLIP-400M & 149M &
  \multicolumn{1}{c}{39.48/40.65} &
  \multicolumn{1}{c}{32.49/32.49} &
  \multicolumn{1}{c}{27.02/27.02} &
  \multicolumn{1}{c}{35.09/44.84} &
  \multicolumn{1}{c}{22.22/26.14} &
  \multicolumn{1}{c}{34.89/42.68} &
  \multicolumn{1}{c}{32.22/43.93} &
  \multicolumn{1}{c}{28.39/29.11} &
                      20.49/20.88 \\

SigLIP~\cite{zhai2023sigmoid} &  WebLI &  149M &
  \multicolumn{1}{c}{46.74/48.34} &
  \multicolumn{1}{c}{41.17/41.17} &
  \multicolumn{1}{c}{31.50/31.50} &
  \multicolumn{1}{c}{45.55/44.78} &
  \multicolumn{1}{c}{22.84/20.20} &
  \multicolumn{1}{c}{50.21/50.12} &
  \multicolumn{1}{c}{47.94/51.89} &
  \multicolumn{1}{c}{44.15/44.13} &
                      \underline{36.11}/\underline{36.19} \\

SigLIP2~\cite{tschannen2025siglip} &  WebLI & 149M &
  \multicolumn{1}{c}{50.37/51.53} &
  \multicolumn{1}{c}{\underline{42.69}/\underline{42.69}} &
  \multicolumn{1}{c}{30.37/30.38} &
  \multicolumn{1}{c}{39.71/40.13} &
  \multicolumn{1}{c}{24.78/27.78} &
  \multicolumn{1}{c}{50.19/\underline{57.57}} &
  \multicolumn{1}{c}{46.17/48.62} &
  \multicolumn{1}{c}{45.60/46.01} &
                      35.93/35.96 \\

EVA-02-CLIP~\cite{sun2023eva}  & Merged-2B & 149M &
  \multicolumn{1}{c}{\underline{52.13}/56.88} &
  \multicolumn{1}{c}{38.43/38.43} &
  \multicolumn{1}{c}{31.57/31.57} &
  \multicolumn{1}{c}{\underline{45.82}/50.52} &
  \multicolumn{1}{c}{\underline{27.14}/27.53} &
  \multicolumn{1}{c}{43.55/43.92} &
  \multicolumn{1}{c}{45.44/42.61} &
  \multicolumn{1}{c}{36.57/37.09} &
                      28.45/28.07 \\

FaRL~\cite{zheng2022general} & LAION-FACE & 149M &
  \multicolumn{1}{c}{30.38/40.74} &
  \multicolumn{1}{c}{28.69/28.69} &
  \multicolumn{1}{c}{25.10/25.10} &
  \multicolumn{1}{c}{28.94/35.10} &
  \multicolumn{1}{c}{13.48/12.12} &
  \multicolumn{1}{c}{32.90/28.78} &
  \multicolumn{1}{c}{38.92/47.91} &
  \multicolumn{1}{c}{30.51/31.46} &
                      22.49/22.74 \\

Exp-CLIP~\cite{zhao2025enhancing} & \tabincell{c}{WIT-400M+CAER-S} & 149M &
  \multicolumn{1}{c}{{48.96/54.50}} & 
  \multicolumn{1}{c}{{39.98/39.98}} &
  \multicolumn{1}{c}{\underline{34.40}/\underline{34.40}} &
  \multicolumn{1}{c}{{40.81/\underline{53.02}}} &
  \multicolumn{1}{c}{19.46/\underline{28.03}} &
  \multicolumn{1}{c}{39.68/41.69} &
  \multicolumn{1}{c}{\underline{53.66}/58.51} &
  \multicolumn{1}{c}{28.33/28.17} &
                     {30.39/32.71} \\

FLIP~\cite{li2024flip}  & \tabincell{c}{WIT-400M+FLIP-80M} & 149M &
  \multicolumn{1}{r}{48.37/\underline{57.17}} &
  \multicolumn{1}{c}{40.69/40.69} &
  \multicolumn{1}{r}{33.85/33.85} &
  \multicolumn{1}{c}{43.05/51.89} &
  \multicolumn{1}{c}{24.35/23.74} &
  \multicolumn{1}{c}{\underline{52.61}/46.65} &
  \multicolumn{1}{c}{51.68/\underline{61.88}} &
  \multicolumn{1}{c}{\underline{53.72}/\underline{53.99}} &
                      31.77/32.48 \\

\rowcolor{color_light_blue}
EmoCapCLIP (Ours) &  \tabincell{c}{WIT-400M+EmoCap-100K} & 149M &
  \multicolumn{1}{c}{\textbf{65.88}/\textbf{68.23}} &
  \multicolumn{1}{c}{\textbf{52.77}/\textbf{52.77}} &
  \multicolumn{1}{c}{\textbf{45.38}/\textbf{45.38}} &
  \multicolumn{1}{c}{\textbf{56.09}/\textbf{66.08}} &
  \multicolumn{1}{c}{\textbf{35.44}/\textbf{42.07}} &
  \multicolumn{1}{c}{\textbf{70.14}/\textbf{68.23}} &
  \multicolumn{1}{c}{\textbf{71.55}/\textbf{73.31}} &
  \multicolumn{1}{c}{\textbf{58.15}/\textbf{58.56}} & 
                       \textbf{48.61}/\textbf{52.60} \\

\midrule
 & \multicolumn{9}{c}{\texttt{CLIP: ViT-L/14}} & \\ 
\midrule

OpenCLIP~\cite{cherti2023reproducible}  & LAION-400M &  428M &
  \multicolumn{1}{c}{41.33/45.83} &
  \multicolumn{1}{c}{34.51/34.51} &
  \multicolumn{1}{c}{27.55/27.55} &
  \multicolumn{1}{c}{39.48/49.87} &
  \multicolumn{1}{c}{20.63/18.69} &
  \multicolumn{1}{c}{42.16/49.88} &
  \multicolumn{1}{c}{42.84/48.42} &
  \multicolumn{1}{c}{31.75/31.92} &
                      26.20/29.93 \\

OpenAI CLIP~\cite{radford2021learning}  & WIT-400M &  428M &
  \multicolumn{1}{c}{49.69/39.18} &
  \multicolumn{1}{c}{38.89/38.89} &
  \multicolumn{1}{c}{33.73/33.73} &
  \multicolumn{1}{c}{44.05/47.80} &
  \multicolumn{1}{c}{26.68/27.15} &
  \multicolumn{1}{c}{54.47/56.08} &
  \multicolumn{1}{c}{64.60/69.01} &
  \multicolumn{1}{c}{29.60/29.11} &
                      35.60/35.03 \\

MetaCLIP~\cite{xudemystifying}  &  MetaCLIP-400M &  428M &
  \multicolumn{1}{c}{44.70/49.97} &
  \multicolumn{1}{c}{38.71/38.71} &
  \multicolumn{1}{c}{32.00/32.00} &
  \multicolumn{1}{c}{41.71/52.31} &
  \multicolumn{1}{c}{22.40/26.52} &
  \multicolumn{1}{c}{48.37/51.12} &
  \multicolumn{1}{c}{41.37/49.64} &
  \multicolumn{1}{c}{24.12/24.88} &
                      29.84/32.02 \\

SigLIP~\cite{zhai2023sigmoid} &  WebLI &  428M &
  \multicolumn{1}{c}{53.49/53.13} &
  \multicolumn{1}{c}{38.40/38.40} &
  \multicolumn{1}{c}{29.43/29.43} &
  \multicolumn{1}{c}{56.12/52.98} &
  \multicolumn{1}{c}{\underline{33.94}/30.18} &
  \multicolumn{1}{c}{51.15/\underline{62.78}} &
  \multicolumn{1}{c}{41.47/43.22} &
  \multicolumn{1}{c}{52.02/52.58} &
                      29.37/32.71 \\

SigLIP2~\cite{tschannen2025siglip} & WebLI &  428M &
  \multicolumn{1}{c}{58.34/62.32} &
  \multicolumn{1}{c}{\underline{45.46}/\underline{45.46}} &
  \multicolumn{1}{c}{38.22/38.22} &
  \multicolumn{1}{c}{\underline{56.93}/50.15} &
  \multicolumn{1}{c}{33.26/32.70} &
  \multicolumn{1}{c}{\underline{59.32}/56.08} &
  \multicolumn{1}{c}{49.08/61.37} &
  \multicolumn{1}{c}{46.72/46.95} &
                      35.89/35.27 \\

EVA-02-CLIP~\cite{sun2023eva}  &  Merged-2B &  428M &
  \multicolumn{1}{c}{58.16/53.72} &
  \multicolumn{1}{c}{40.77/40.77} &
  \multicolumn{1}{c}{32.77/32.77} &
  \multicolumn{1}{c}{53.68/48.87} &
  \multicolumn{1}{c}{33.73/33.08} &
  \multicolumn{1}{c}{52.94/54.59} &
  \multicolumn{1}{c}{50.27/55.05} &
  \multicolumn{1}{c}{49.12/49.30} &
                      32.80/34.34 \\

Exp-CLIP~\cite{zhao2025enhancing} &  \tabincell{c}{WIT-400M+CAER-S} &  428M &
  \multicolumn{1}{c}{\underline{58.70}/\underline{65.37}} & 
  \multicolumn{1}{c}{{44.27/44.27}} &
  \multicolumn{1}{c}{\underline{38.44}/\underline{38.43}} &
  \multicolumn{1}{c}{{48.28/\underline{55.42}}} &
  \multicolumn{1}{c}{26.23/32.07} &
  \multicolumn{1}{c}{56.95/52.36} &
  \multicolumn{1}{c}{\underline{74.77}/\underline{81.14}} &
  \multicolumn{1}{c}{\underline{55.76}/\underline{55.40}} &
                     {31.93/34.34} \\

FLIP~\cite{li2024flip}   &   \tabincell{c}{WIT-400M+FLIP-80M} &   428M &
  \multicolumn{1}{r}{52.03/56.29} &
  \multicolumn{1}{c}{42.86/42.86} &
  \multicolumn{1}{r}{35.83/35.83} &
  \multicolumn{1}{c}{45.92/46.32} &
  \multicolumn{1}{c}{29.93/\underline{39.27}} &
  \multicolumn{1}{c}{54.20/54.09} &
  \multicolumn{1}{c}{57.21/61.57} &
  \multicolumn{1}{c}{44.66/45.07} &
                      \underline{36.18}/\underline{38.05} \\

\rowcolor{color_light_blue}
EmoCapCLIP (Ours) & \tabincell{c}{WIT-400M+EmoCap-100K} &   428M &
  \multicolumn{1}{c}{\textbf{68.91}/\textbf{70.45}} &
  \multicolumn{1}{c}{\textbf{53.37}/\textbf{53.37}} &
  \multicolumn{1}{c}{\textbf{46.50}/\textbf{46.50}} &
  \multicolumn{1}{c}{\textbf{63.08}/\textbf{62.02}} &
  \multicolumn{1}{c}{\textbf{41.40}/\textbf{45.10}} &
  \multicolumn{1}{c}{\textbf{73.15}/\textbf{70.95}} &
  \multicolumn{1}{c}{\textbf{76.33}/\textbf{82.06}} &
  \multicolumn{1}{c}{\textbf{57.66}/\textbf{57.55}} & 
                      \textbf{46.56}/\textbf{49.22} \\

\bottomrule
\end{tabular}
}
\end{center}
\end{table*}
\begin{table*}[htbp]
\scriptsize
\begin{center}
\caption{Zero-shot performance \textit{vs.} cross-dataset performance of traditional supervised methods.}
\label{tab:zero_shot_vs_supervised_cross_eval}
\resizebox{\linewidth}{!}{
\setlength{\tabcolsep}{1pt} 
\begin{tabular}{r*{19}{c}c} 
\toprule
\rotatebox{60}{\textbf{Methods}} & \rotatebox{60}{SCN~\cite{wang2020suppressing}} & \rotatebox{60}{RUL~\cite{zhang2021relative}} & \rotatebox{60}{EAC~\cite{zhang2022learn}} & \rotatebox{60}{OFER~\cite{lee2023latent}} & \rotatebox{60}{CAFE~\cite{zhang2024generalizable}} 
& \rotatebox{60}{SCN~\cite{wang2020suppressing}} & \rotatebox{60}{RUL~\cite{zhang2021relative}} & \rotatebox{60}{EAC~\cite{zhang2022learn}} & \rotatebox{60}{OFER~\cite{lee2023latent}} & \rotatebox{60}{CAFE~\cite{zhang2024generalizable}} 
 & \rotatebox{60}{SCN~\cite{wang2020suppressing}} & \rotatebox{60}{RUL~\cite{zhang2021relative}} & \rotatebox{60}{EAC~\cite{zhang2022learn}} & \rotatebox{60}{OFER~\cite{lee2023latent}} & \rotatebox{60}{\tabincell{c}{CLIP~\cite{radford2021learning}}} & \rotatebox{60}{CAFE~\cite{zhang2024generalizable}}  & \rotatebox{60}{S2D~\cite{chen2024static}} & \rotatebox{60}{FaRL~\cite{zheng2022general}} & \rotatebox{60}{\tabincell{c}{ExpLLM~\cite{lan2025expllm}}} 
& \cellcolor{color_light_blue} \rotatebox{60}{\tabincell{c}{EmoCapCLIP\\(Ours)}}
\\
\cmidrule(lr){2-6} \cmidrule(lr){7-11} \cmidrule(lr){12-20}
 & \multicolumn{5}{c}{\textbf{\tabincell{c}{Train on AffectNet-7}}} & \multicolumn{5}{c}{\textbf{\tabincell{c}{Train on SFEW2.0}}}  & \multicolumn{9}{c}{\textbf{\tabincell{c}{Train on RAF-DB}}} &  \textbf{Zero-Shot} \\

\midrule

\rotatebox{0}{\textbf{\tabincell{r}{Test on\\AffectNet-7}}} & - & - & - & - & - &   23.19  & 30.41 & 25.57 & 23.02 & 32.24     & 42.85 & 43.82 & 43.91 & 42.73 & 44.31 & 45.86  & 47.57   & 49.33 & \underline{51.86}   & 
\cellcolor{color_light_blue} \textbf{51.91} 
\\

\rotatebox{0}{\textbf{\tabincell{r}{Test on\\SFEW2.0}}}     & 41.98 & 34.01 & 44.89 & 42.60 & 51.18   & - & - & - & - & -     & 44.89 & 46.91 & 43.39 & 43.88 & 41.30 & \underline{52.86}   & -  & - & -           & 
\cellcolor{color_light_blue} \textbf{53.06}
\\

\rotatebox{0}{\textbf{\tabincell{r}{Test on\\RAF-DB}}}      & \underline{70.70} & 55.83 & 66.10 & 63.21 & \textbf{72.69}   & 43.10 & 45.90 & 45.79 & 44.52 & 49.38  & - & - & - & - & - & - & -               & - & -        & 
\cellcolor{color_light_blue} 67.03
\\

\bottomrule
\end{tabular}

}
\end{center}
\end{table*}

\begin{table*}[!t]
\scriptsize
\begin{center}
\caption{
Zero-shot DFER (UAR/WAR) results in comparison with SOTA methods using ViT-B/32.  
$^\dag$: video CLIP models with temporal information modeling. $^\diamondsuit$: use human-annotated captions.
}
\label{tab:zs_dfer_vit-b-32}
\resizebox{\linewidth}{!}{
\setlength{\tabcolsep}{4pt} 
\begin{tabular}{lccccrccc}
\toprule

\textbf{Methods} &
  \multicolumn{1}{c}{\textbf{DFEW}} &
  \multicolumn{1}{c}{\textbf{FERV39k}} &
  \multicolumn{1}{c}{\textbf{MAFW}} &
  \multicolumn{1}{c}{\textbf{AFEW}} &
  \multicolumn{1}{c}{\textbf{\tabincell{c}{MAFW\\(Compound)}}} &
  \multicolumn{1}{c}{\textbf{CREMA-D}} &
  \multicolumn{1}{c}{\textbf{RAVDESS}} &
  \multicolumn{1}{c}{\textbf{eNTERFACE05}}
                      \\

\midrule

OpenCLIP~\cite{cherti2023reproducible} &
  \multicolumn{1}{c}{20.04/19.23} &
  \multicolumn{1}{c}{19.48/17.45} &
  \multicolumn{1}{c}{12.95/12.76} &
  \multicolumn{1}{c}{18.30/16.45} &
  \multicolumn{1}{c}{3.35/6.09} &
  \multicolumn{1}{c}{23.16/22.99} &
  \multicolumn{1}{c}{18.23/19.38} &
  \multicolumn{1}{c}{18.94/18.96}
  \\

OpenAI CLIP~\cite{radford2021learning}  &
  \multicolumn{1}{c}{22.08/21.30} &
  \multicolumn{1}{c}{18.12/17.38} &
  \multicolumn{1}{c}{13.39/15.48} &
  \multicolumn{1}{c}{28.59/28.87} &
  \multicolumn{1}{c}{5.36/\underline{11.39}} &
  \multicolumn{1}{c}{24.96/23.77} &
  \multicolumn{1}{c}{22.40/17.22} &
  \multicolumn{1}{c}{20.94/20.98}
  \\

MetaCLIP~\cite{xudemystifying} &
  \multicolumn{1}{c}{21.73/26.01} &
  \multicolumn{1}{c}{19.68/24.46} &
  \multicolumn{1}{c}{14.59/10.02} &
  \multicolumn{1}{c}{20.05/20.10} &
  \multicolumn{1}{c}{3.97/4.86} &
  \multicolumn{1}{c}{19.30/19.16} &
  \multicolumn{1}{c}{21.68/17.50} &
  \multicolumn{1}{c}{18.31/18.34}
  \\

SigLIP2~\cite{tschannen2025siglip} &
  \multicolumn{1}{c}{30.89/24.63} &
  \multicolumn{1}{c}{21.96/19.68} &
  \multicolumn{1}{c}{19.46/16.16} &
  \multicolumn{1}{c}{27.66/26.11} &
  \multicolumn{1}{c}{5.00/7.97} &
  \multicolumn{1}{c}{31.63/31.31} &
  \multicolumn{1}{c}{\underline{36.65}/\underline{32.43}} &
  \multicolumn{1}{c}{23.78/23.78}
                      \\

Exp-CLIP~\cite{zhao2025enhancing} &
  \multicolumn{1}{c}{24.25/25.87} &
  \multicolumn{1}{c}{21.48/21.25} &
  \multicolumn{1}{c}{17.53/20.27} &
  \multicolumn{1}{c}{29.72/31.23} &
  \multicolumn{1}{c}{4.42/10.06} &
  \multicolumn{1}{c}{24.98/24.50} &
  \multicolumn{1}{c}{21.22/17.43} &
  \multicolumn{1}{c}{26.45/26.42}
                      \\

FLIP~\cite{li2024flip} &
  \multicolumn{1}{c}{27.46/25.11} &
  \multicolumn{1}{c}{20.21/15.69} &
  \multicolumn{1}{c}{20.93/22.99} &
  \multicolumn{1}{c}{21.63/20.63} &
  \multicolumn{1}{c}{6.00/9.26} &
  \multicolumn{1}{c}{24.97/25.36} &
  \multicolumn{1}{c}{26.17/25.83} &
  \multicolumn{1}{c}{20.42/20.44}
  \\

VideoCLIP$^\dag$~\cite{xu2021videoclip} &
  \multicolumn{1}{c}{18.15/17.52} &
  \multicolumn{1}{c}{15.35/15.57} &
  \multicolumn{1}{c}{10.26/11.70} &
  \multicolumn{1}{c}{16.32/17.32} &
  \multicolumn{1}{c}{-} &
  \multicolumn{1}{c}{-} &
  \multicolumn{1}{c}{-} &
  \multicolumn{1}{c}{-}
                      \\
X-CLIP$^\dag$~\cite{ni2022expanding}  &
  \multicolumn{1}{c}{20.03/16.06} &
  \multicolumn{1}{c}{16.80/12.50} &
  \multicolumn{1}{c}{11.51/13.53} &
  \multicolumn{1}{c}{21.81/22.05} &
  \multicolumn{1}{c}{-} &
  \multicolumn{1}{c}{-} &
  \multicolumn{1}{c}{-} &
  \multicolumn{1}{c}{-}

  \\

EmotionCLIP$^\dag$~\cite{zhang2023learning} &
  \multicolumn{1}{c}{13.77/19.89} &
  \multicolumn{1}{c}{14.79/19.54} &
  \multicolumn{1}{c}{9.20/11.65} &
  \multicolumn{1}{c}{14.86/17.06} &
  \multicolumn{1}{c}{-} &
  \multicolumn{1}{c}{-} &
  \multicolumn{1}{c}{-} &
  \multicolumn{1}{c}{-}
  \\

EmoCLIP$^\dag$$^\diamondsuit$~\cite{foteinopoulou2024emoclip} &
  \multicolumn{1}{c}{\underline{36.76}/\textbf{46.27}} &
  \multicolumn{1}{c}{\underline{26.73}/\textbf{35.30}} &
  \multicolumn{1}{c}{\underline{25.86}/\underline{33.49}} &
  \multicolumn{1}{c}{\underline{36.13}/\underline{39.90}} &
  \multicolumn{1}{c}{\underline{6.58}/\textbf{18.53}}  &
  \multicolumn{1}{c}{\underline{33.13}/\underline{31.91}}  &
  \multicolumn{1}{c}{31.51/30.21}  &
  \multicolumn{1}{c}{\underline{28.06}/\underline{27.89}}
                     \\

\rowcolor{color_light_blue}
  EmoCapCLIP (Ours) &
  \multicolumn{1}{c}{\textbf{42.19}/\underline{43.99}} &
  \multicolumn{1}{c}{\textbf{29.87}/\underline{31.99}} &
  \multicolumn{1}{c}{\textbf{30.85}/\textbf{34.50}} &
  \multicolumn{1}{c}{\textbf{39.94}/\textbf{40.30}} &
  \multicolumn{1}{c}{\textbf{9.31}/9.93} &
  \multicolumn{1}{c}{\textbf{48.80}/\textbf{47.99}} &
  \multicolumn{1}{c}{\textbf{50.43}/\textbf{50.86}} &
  \multicolumn{1}{c}{\textbf{35.26}/\textbf{35.14}}
\\

\bottomrule
\end{tabular}
}
\vspace{-20pt}
\end{center}
\end{table*}

\textbf{Zero-shot DFER.}
To investigate the transferability of the learned representations in the video domain, we further evaluate EmoCapCLIP on 8 DFER benchmarks. 
For image CLIP models, we sample 16 frames per video and apply temporal pooling to obtain the video representation~\cite{zhao2025enhancing}.
Besides, the same prompt as SFER is used for evaluation. 
As shown in Table~\ref{tab:zs_dfer_vit-b-32}, our method achieves much better performance than all image CLIP baselines. 
Compared to general-purpose and domain-specific video CLIP models with complex temporal information modeling, EmoCapCLIP also shows competitive or superior performance on all benchmarks. 
Notably, although EmoCLIP~\cite{foteinopoulou2024emoclip} utilizes human-annotated captions from MAFW for supervision, it still lags behind EmoCapCLIP in terms of UAR on several in-the-wild benchmarks (e.g., 25.86\% \textit{vs.} 30.85\% on MAFW) and lab-controlled benchmarks (e.g., 33.13\% \textit{vs.} 48.80\% on CREMA-D). 
This finding suggests that large-scale, semantically rich captions from MLLMs can be effectively leveraged within a well-designed contrastive learning framework to learn transferable facial emotion representations, without relying on costly manual annotations.

\begin{figure*}[htbp]
    \centering
    \includegraphics[trim={0 0.6cm 0 0},clip,width=0.96\linewidth]{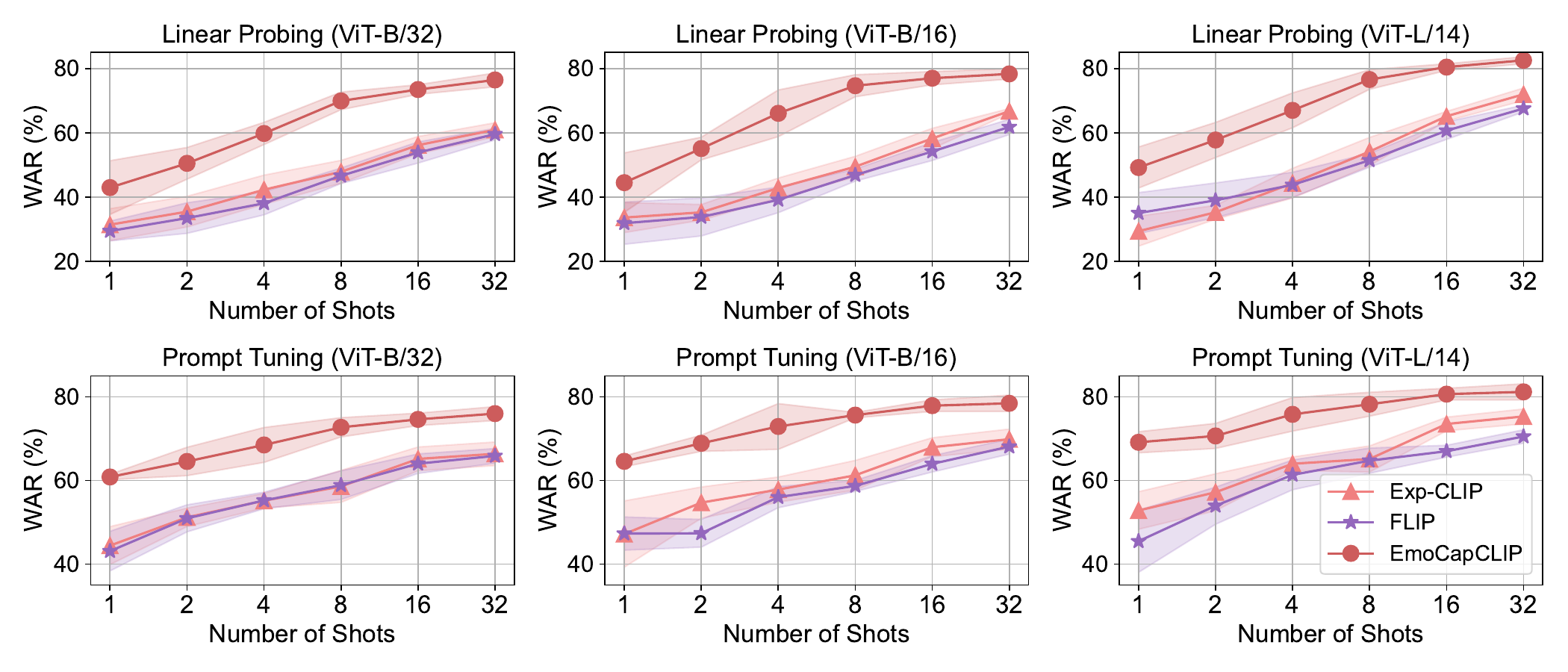}
    \caption{Few-shot linear probing (Upper) and prompt tuning (Bottom) results on RAF-DB.}
    \label{fig:fs_sfer}
\end{figure*}
\begin{figure*}[htbp]
    \centering
    \includegraphics[trim={0 0.5cm 0 0},clip,width=\linewidth]{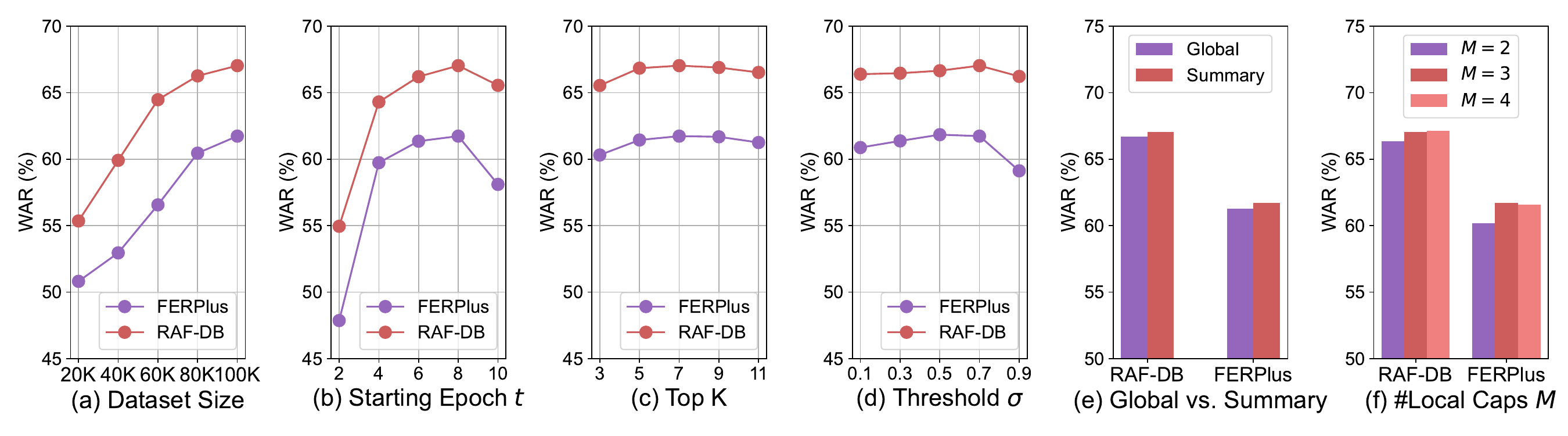}
    \caption{Ablation studies on miscellaneous design choices.}
    \label{fig:ablation_misc}
\end{figure*}

\textbf{Few-shot SFER.} 
The above parts focus on the zero-shot evaluation of the learned representations. We are also interested in the adaptation of EmoCapCLIP within the few-shot setting. 
To investigate this, we follow CoOp~\cite{zhou2022learning} to employ two prominent few-shot adaptation methods: linear probing and prompt tuning.
The results on RAF-DB are shown in Figure~\ref{fig:fs_sfer}. Each experiment was run five times with different seeds to ensure the reliability of the evaluation. As can be seen in the figure, EmoCapCLIP significantly outperforms Exp-CLIP and FLIP across different backbones and evaluation methods.
Notably, this performance advantage is more pronounced when the number of shots is smaller, demonstrating the superior quality and high transferability of the representations learned by EmoCapCLIP, particularly in low-data regimes.

\subsection{Ablation Study}
\label{sec:exp_ablation}

In this study, we compare the zero-shot SFER performance on RAF-DB and FERPlus using ViT-B/32.

\setlength{\intextsep}{-7pt}
\begin{wraptable}[9]{r}{0.6\textwidth}  
\scriptsize
\begin{center}
\caption{Ablation study of main components.}
\label{tab:ablation_main}
\begin{tabular}{cccccc}
\toprule
$\mathcal{L}_g$ & $\mathcal{L}_r^{\textrm{intra}}$  & $\mathcal{L}_r^{\textrm{inter}}$   & CMGPM  & RAF-DB & FERPlus  \\
\midrule
\cmark    & \xmark   & \xmark  & \xmark    &  60.41/63.85 & 52.26/57.09        \\
\cmark    & \cmark   & \xmark  & \xmark    &  61.02/64.48 & 52.82/58.41        \\
\cmark    & \xmark   & \cmark  & \xmark    &  61.18/64.53 & 53.04/58.17        \\
\cmark    & \cmark   & \cmark  & \xmark    &  61.59/65.39 & 53.79/58.87        \\
\cmark    & \cmark   & \xmark  & \cmark    &  62.17/66.28 & 54.10/59.93        \\
\cmark    & \xmark   & \cmark  & \cmark    &  62.21/66.79 & 54.50/60.62        \\
\cmark    & \cmark   & \cmark  & \cmark    &  \textbf{62.89}/\textbf{67.03} & \textbf{55.10}/\textbf{61.73}        \\
\bottomrule
\end{tabular}
\end{center}
\end{wraptable}

\textbf{Effectiveness of Main Components.}
We first ablate several main components in EmoCapCLIP, including the global contrastive loss $\mathcal{L}_g$, the intra- and inter-sample local contrastive loss ($\mathcal{L}^{\textrm{intra}}_r$ and $\mathcal{L}^{\textrm{inter}}_r$), and the cross-modal guided positive mining (CMGPM) module. 
As shown in Table~\ref{tab:ablation_main}, we start with the simple baseline using only $\mathcal{L}_g$. It achieves reasonable performance on two datasets, which serves as a good starting point. Subsequently, the incorporation of $\mathcal{L}^{\textrm{intra}}_r$ or $\mathcal{L}^{\textrm{inter}}_r$ boosts the performance over the baseline. Besides, combining them yields additional gains. 
These results indicate that the local contrastive learning enforced by $\mathcal{L}^{\textrm{intra}}_r$ and $\mathcal{L}^{\textrm{inter}}_r$ effectively supplements the global contrastive learning, facilitating a more comprehensive and nuanced representation of facial emotions.
Furthermore, the inclusion of the CMGPM module yields consistent performance across different configurations. This outcome underscores the critical importance of mining high-quality positive samples with similar emotional states to enhance the contrastive learning process.
The final configuration integrating all components achieves the highest scores, validating the contribution of each element to the effectiveness of EmoCapCLIP.

\textbf{Dataset Size.} 
We investigate how the amount of training data affects the quality of the learned representations. 
As shown in Figure~\ref{fig:ablation_misc}(a), the model performance is limited when only a small amount of data is available. However, substantial performance improvement is observed as the data volume increases, culminating in the best results achieved at the full 100K scale. This finding highlights the critical importance of sufficient data volume for learning strong emotion-aware facial representations from semantically rich captions.

\textbf{Several Factors in CMGPM.} 
We then investigate the influence of the activation timing of the CMGPM mechanism, the similarity threshold $\sigma$, and the number of top $K$ similar samples used for positive set construction. 
The results shown in Figure~\ref{fig:ablation_misc}(b) first reveal that a suitable activation timing for CMGPM is crucial to achieve good performance. Initiating CMGPM prematurely, during the early stage of model training, is detrimental. This is intuitive, as the representation space is not yet sufficiently well-formed, rendering the positive samples mined by CMGPM at this stage less reliable. Conversely, delaying CMGPM activation until the final epoch is also suboptimal, limiting its potential impact on the learning process.
Regarding the ablation of $K$ and $\sigma$ shown in Figure~\ref{fig:ablation_misc}(c) and (d), we find that the model performance shows relatively low sensitivity to their specific choices.

\textbf{Caption Selection for $\mathcal{L}_g$.}
There are two caption choices for global contrastive learning, i.e., the global part ($T^i_{\texttt{\texttt{Global}}}$) and the summary part ($T^i_{\texttt{\texttt{Summary}}}$) in the full caption. Figure~\ref{fig:ablation_misc}(e) shows that employing the latter results in slightly superior performance.

\textbf{Number of Local Captions $M$.}
We finally ablate three different choices of $M$ for local contrastive learning in Figure~\ref{fig:ablation_misc}(f). Results show that larger $M$ generally leads to slight performance improvement. Since larger $M$ incurs greater GPU memory demands, we adopt $M=3$ by default.

\section{Conclusion}

In this work, we have embraced large-scale semantically rich captions to unleash the power of natural language supervision for facial emotion representation learning. 
To overcome the data bottleneck, we established EmoCap100K, a large-scale facial emotion caption dataset featuring comprehensive and structured affective descriptions. We also proposed EmoCapCLIP, a novel framework tailored to fully harness the rich supervisory signals in captions and facilitate better emotion understanding.
Extensive evaluations on diverse benchmarks and tasks demonstrate the strong transferability of the learned representations and open new research avenues for emotion-centric vision-language pre-training.



{
\small
\bibliographystyle{unsrtnat}
\bibliography{main}
}

\end{document}